\documentclass[sigconf]{acmart}
\AtBeginDocument{%
  \providecommand\BibTeX{{%
    \normalfont B\kern-0.5em{\scshape i\kern-0.25em b}\kern-0.8em\TeX}}}
\setcopyright{none}
\renewcommand\footnotetextcopyrightpermission[1]{} 
\newcommand\blfootnote[1]{%
  \begingroup
  \renewcommand\thefootnote{}\footnote{#1}%
  \addtocounter{footnote}{-1}%
  \endgroup
}
\usepackage{amsmath,amsfonts}
\usepackage{algorithm}
\usepackage{algorithmic}
\usepackage{graphicx}
\usepackage{textcomp}
\usepackage{xcolor}
\usepackage{microtype}
\usepackage{multirow}
\usepackage{subfigure}
\usepackage{booktabs} 
\usepackage{hyperref} 
\hypersetup{
    colorlinks=true,
    linkcolor=black,
    citecolor = black,
    urlcolor=black,}
\usepackage{xspace}
\usepackage{diagbox}
\usepackage{array}
 \usepackage{float}
\def\cA{\mathcal{A}}
\def\cD{\mathcal{D}}
\def\cF{\mathcal{F}}
\def\cG{\mathcal{G}}
\def\cH{\mathcal{H}}
\def\cI{\mathcal{I}}
 
\def\cL{\mathcal{L}}

\def\bbX{{\mathbb X}}

\def\tH{{\mathtt H}}
\def\tI{{\mathtt I}}

\def\bS{{\mathbf{S}}}
\def\bx{{\mathbf x}}
\def\bX{{\mathbf X}}

\def\bY{{\mathbf Y}}
\def\bz{{\mathbf z}}

\def\sb{\overline{s}}
\def\yb{\overline{y}}
\def\sh{\hat{{s}}}
\def\yh{\hat{{y}}}
\def\bsh{\hat{{\mathbf{s}}}}
\def\byh{\hat{{\mathbf{y}}}}
\def\bSh{\hat{{\mathbf{S}}}}
\def\bYh{\hat{{\mathbf{Y}}}}

\def\bbD{{\mathbb{D}}}
\def\bbF{{\mathbb{F}}}
\def\bbP{{\mathbb P}}
\def\bbR{{\mathbb{R}}}
\def\bbI{{\mathbb{I}}}

\def\dhon{\mbox{$\delta^y$-$honest$ }}
\def\dcur{\mbox{$\delta^s$-$curious$ }}

\def\sfR{\mathsf{R}}
\def\sfP{\mathsf{P}}
\def\NC{$\mathsf{Std}$\xspace} 
\def\HBCR{$\mathsf{Raw{HBC}}$\xspace} 
\def\HBCP{$\mathsf{Soft{HBC}}$\xspace}

\newcommand{\latinphrase}[1]{\textit{#1}}

\newcommand{\ie}{\latinphrase{i.e.,}\xspace}
\newcommand{\eg}{\latinphrase{e.g.,}\xspace}
\newcommand{\wrt}{\latinphrase{w.r.t.}\xspace}
\newcommand{\aka}{\latinphrase{a.k.a.}\xspace}
\newcommand{\pub}{{target}\xspace}
\newcommand{\priv}{{sensitive}\xspace}
\newcommand{\user}{user\xspace}
\newcommand{\model}{classifier\xspace}
\newcommand{\Model}{Classifier\xspace}
\newcommand{\Models}{Classifiers\xspace}
\renewcommand{\models}{classifiers\xspace}
\newcommand{\users}{users\xspace}

\newcommand{\Users}{Users\xspace}
\newcommand{\server}{server\xspace}
\newcommand{\servers}{servers\xspace}

\newcommand{\hbc}{HBC\xspace}
\newcommand{\nc}{standard\xspace}
\newcommand{\Nc}{Standard\xspace}
\newcommand{\Rval}{raw\xspace}
\newcommand{\Pval}{soft\xspace}
\newcommand{\round}[1]{\lfloor{#1}\rceil}  
\begin{document}
\title{Honest-but-Curious Nets: Sensitive Attributes of Private Inputs Can Be Secretly Coded into the Classifiers' Outputs} 

\author{Mohammad Malekzadeh}
\affiliation{\country{Imperial College London, UK}}
\email{m.malekzadeh@imperial.ac.uk}
\author{Anastasia Borovykh}
\affiliation{\country{Imperial College London, UK}}
\email{a.borovykh@imperial.ac.uk}
\author{Deniz Gündüz}
\affiliation{\country{Imperial College London, UK}}
\email{d.gunduz@imperial.ac.uk}
\fancyhead{}
\fancyhf{}
\fancyfoot[C]{\thepage}

\begin{abstract}
It is known that deep neural networks, trained for the classification of non-sensitive {\em \pub} attributes, can reveal  {\em \priv} attributes of their input data through internal representations extracted by the \model. We take a step forward and show that deep \models can be trained to secretly encode a \priv attribute of their input data into the \model's outputs for the \pub attribute, at inference time. Our proposed attack works even if users have a full white-box view of the \model, can keep all internal representations hidden, and only release the \model's estimations for the \pub attribute. We introduce an information-theoretical formulation for such attacks and present efficient empirical implementations for training {\em honest-but-curious~(\hbc)} \models: {\em \models that can be accurate in predicting their \pub attribute, but can also exploit their outputs to secretly encode a \priv attribute}. Our work highlights a vulnerability that can be exploited by malicious machine learning service providers to attack their user's privacy in several seemingly safe scenarios; such as encrypted inferences, computations at the edge, or private knowledge distillation. Experimental results on several attributes in two face-image datasets show that a semi-trusted \server can train \models that are not only perfectly {\em honest} but also accurately {\em curious}. We conclude by showing the difficulties in distinguishing between \nc and \hbc \models, discussing challenges in defending against this vulnerability of deep classifiers, and enumerating related open
directions for future studies.  
\vspace{.7cm}
\end{abstract}


\vspace{.7cm}


\maketitle
\vspace{.7cm}
\section{Introduction}
\begin{figure}[t]
    \centering
    \vspace{.5cm}
    \includegraphics[width=\columnwidth]{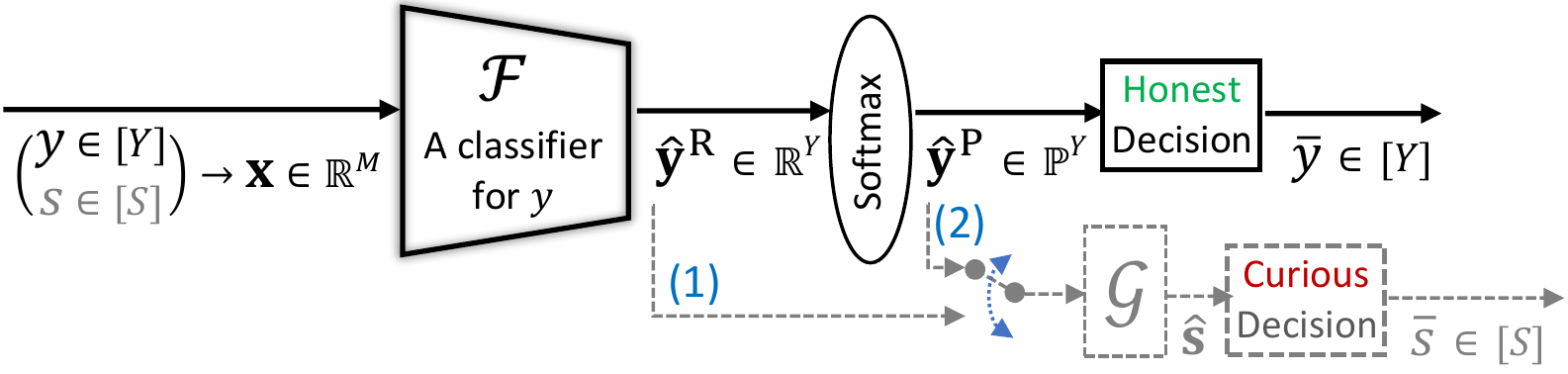}
    \caption{Let {\em \user}s' data $\bx$~(\eg face image) contain a {\em \pub} attribute~$y$ (\eg Age) and a {\em \priv} attribute~$s$ (\eg Race). A {\em \model} $\cF$ is provided by a {\em \server} to estimates $y$, and \users only release the \model's outputs $\byh$.  We show that $\cF$ can be trained such that $\byh$ is not only accurate for $y$~({\em honesty}), but can also be used by a secret attack $\cG$ to infer $s$~({\em curiosity}). We present efficient attacks in two scenarios where either (1)~the {\em \Rval} outputs  $\byh^{\sfR}$, or (2) the {\em \Pval} outputs $\byh^{\sfP}$ are released.\label{fig:classifier}}
\end{figure} 
Machine learning~(ML)\blfootnote{Will be appeared in ACM Conference on Computer and Communications Security (CCS '21), November 15–19, 2021.} \models, trained on a set of labeled data, aim to facilitate the estimation of a {\em \pub} label (\ie~{\em attribute}) for new data at inference time; from smile detection for photography~\cite{whitehill2009toward} to automated detection of a disease on medical data~\cite{gulshan2016development, esteva2017dermatologist}. However, in addition to the \pub attribute that the \model is trained for, data might also contain some other {\em \priv} attributes. For example, there are several attributes that can be inferred from a face image; such as gender, age, race, emotion, hairstyle, and more~\cite{liu2015celeba, zhang2017age}. Since ML \models, particularly  deep neural networks~(DNNs), are becoming increasingly popular, either as cloud-based services or as part of apps on our personal devices, it is important to be aware of the type of \priv attributes that we might reveal through using these \models; especially when a \model is supposed to only estimate a specified attribute. For instance, while clinical experts can barely identify the race of patients from their medical images, DNNs show considerable performance in detecting race from chest X-rays and CT scans~\cite{banerjee2021reading}.  

In two-party computations, a legitimate party that does not deviate from its specified protocol but attempts to infer as much \priv information as possible from the received data is called a {\em honest-but-curious}~(\hbc) party~\cite{chor1991zero, kushilevitz1994reducibility, goldreich2009foundations, paverd2014modelling}. Following convention, if a {\em \model's outputs} not only allow to estimate the \pub attribute, but also reveal information about other attributes (particularly those uncorrelated to the \pub one) we call it an {\em \hbc \model}.  In this paper, we show how a semi-trusted \server can train an \hbc \model such that the outputs of the \model are not only useful for inferring the \pub attribute, but can also secretly carry information about a \priv attribute of the \user's data that is unrelated to the \pub attribute. 

Figure~\textcolor{blue}{\ref{fig:classifier}} shows an overview of the problem. We consider a {\em \server} that  provides its {\em \users} access to a {\model} trained for a known \pub attribute $y$. We put no restriction on the input data $\bx$ or the \users' access to the \model: \users can have control over the \model and get a white-box view to it~(\eg when the \model is deployed on the \users' devices), or \users can perform secure computation on their data if they have a black-box view~(\eg when the \model is hosted in the cloud). Our only assumption is that both the \user and the \server can observe the outputs of \model $\byh$. For a $Y$-class \model, $\byh$ is a real-valued vector of size $Y$ containing either (1)~{\em \Rval} scores  $\byh^{\sfR} \in \bbR^{Y}$, or (2) {\em \Pval} scores $\byh^{\sfP} = \texttt{softmax}(\byh^{\sfR}) \in \bbP^{Y}$, where the $i$-th \Pval score is $\byh^{\sfP}_i = e^{\byh^{\sfR}_i}/\sum_{j=0}^{Y-1} e^{\byh^{\sfR}_j}$. The \server usually uses some threshold functions to decide $\yb$ as the final predicted class. In classification tasks, an estimated probability distribution over possible classes is more useful than just receiving the most probable class; as it allows the aggregation of outputs provided by multiple ML services to enhance the ultimate decision. Moreover, collecting outputs can help a \server to monitor and enhance its decisions and the provided services.

To protect the {\em user}'s privacy, it is usually proposed to hide the data, as well as all the intermediate computations on the data, and only release the \model's outputs; either using secure two-party computation via cryptography~\cite{beaver1991perfect, gentry2013homomorphic,agrawal2019quotient} or by restricting the computations to edge devices~\cite{teerapittayanon2017distributed, mo2020darknetz}. Although these encrypted or edge solutions hide the input data and the internal features extracted by the \model, the outputs are usually released to the service provider, because the estimation of a \pub attribute might not seem sensitive to the \user's privacy and it might be needed for further services offered to the user. For instance, an insurance company raised huge ethical concerns, when it announced that its ML model extracts ``non-verbal cues'' from videos of users' faces to identify fraud~\cite{lemonade2021tweet}. We show that even if the video is processed locally, or in an encrypted manner, and only a single real-valued output $\yh \in [0,1]$ is released to the insurance company, as the probability of fraud, this single output can still be designed to reveal another sensitive attribute about the \user. In our experiments, we show that a $\yh \in [0,1]$ produced by a smile-detection \model, as the probability of ``smiling'', can  be used to secretly infer whether that person is ``white'' or not\footnote{See Appendix~\textcolor{blue}{\ref{sec:more_motiv_example}}
for more motivational examples of \hbc \models that can be trained for other types of \users' data, such as text, motion sensors, and audio.}.  

We first show that in a {\em black-box} view, where an arbitrary architecture can be used for the \model, the \server can obtain the best achievable trade-off between {\em honesty}~(\ie the classification accuracy for the \pub attribute) and {\em curiosity}~(\ie the classification accuracy for the \priv attribute).  Specifically, we build a controlled synthetic dataset and show how to create such an \hbc \model via a weighted mixture of two separately trained \models, one for the \pub attribute and another for the \priv  attribute~(Section~\textcolor{blue}{\ref{sec:black-box}}). Then, we focus on the more challenging, {\em white-box} view where the \server might have some constraints on the chosen model, \eg the restriction to not being suspicious, or that the \model must be one of the known off-the-shelf models~(Section~\textcolor{blue}{\ref{sec:white-box}}). To this end, we formulate the problem of training an \hbc \model in a general information-theoretical framework, via the {\em information bottleneck} principle~\cite{tishby2000information}, and show the existence of a general  attack  for encoding a desired \priv data into the output of a \model. We propose two practical methods that can be used by a \server for building an \hbc \model, one via the {\em regularization} of \model's loss function, and another via training of a {\em parameterized} model.

Extensive experiments, using typical DNNs for several tasks with different attributes defined on two real-world datasets~\cite{liu2015celeba, zhang2017age}, show that \hbc \models can mostly achieve honesty very close to \nc \models, while also being very successful in their curiosity~(Section~\textcolor{blue}{\ref{sec:eval}}). We, theoretically and empirically, show that the entropy of an \hbc \model's outputs usually tends to be higher than the entropy of a \nc \model's outputs. Moreover, we explain how a \server can improve the honesty of the \model by trading some curiosity via adding an entropy minimization component to parameterized attacks, which in particular, can make \hbc \models less suspicious against proactive defenses~\cite{kearns2020ethical}.
 
Previous works propose several types of attacks {\em to} ML models~\cite{mirshghallah2020privacy, de2020overview}, mostly to DNNs~\cite{liu2021ml}, including property inference~\cite{melis2019exploiting}, membership inference~\cite{shokri2017membership,salem2018ml}, model inversion~\cite{fredrikson2015model}, model extraction~\cite{tramer2016stealing, jagielski2020high}, adversarial examples~\cite{szegedy2013intriguing}, or model poisoning~\cite{biggio2012poisoning, jagielski2018manipulating}. But these attacks mostly concern the privacy of the training dataset and, in all these attacks, the ML model is the trusted party, while \users are assumed untrusted. Our work, from a different point of view, discusses a new threat model, where (the owner of) the ML model is semi-trusted and might attack the privacy of its users at inference time. The closest related work is  the {\em ``overlearning''} concept in~\cite{song2019overlearning}, where it is shown that internal representations extracted by DNN layers can reveal \priv attributes of the input data that might not even be correlated to the \pub attribute. The assumption of~\cite{song2019overlearning} is that an adversary observes a subset of internal representations, while we assume all internal representations to be hidden and an adversary has access only to the outputs. Notably, we show that when \users only release the \model's outputs, overlearning is not a major concern as \nc \models do not reveal significant information about a \priv attribute through their outputs, whereas an \hbc \model can secretly, and almost perfectly, reveal a \priv attribute just via \model's outputs. \\

\textbf{Contributions.} In summary, this paper proposes the following contributions to advance privacy protection in using ML services. 

(1)~We show that ML services can attack their \user's privacy even in a highly restricted setting where they  can only get access to the results of an agreed \pub computation on their \users' data. 

(2)~We formulate such an attack in a general information-theoretical formulation and show the efficiency of our attack via several empirical results.  Mainly, we show how \hbc \models can encode a \priv attribute of their private input into the \model's output, by exploiting the output's entropy as a side-channel. Therefore, \hbc \models tend to produce higher-entropy outputs than \nc \models. However, we also show that the output's entropy can be efficiently reduced by trading a small amount of curiosity of the \model, thus making it even harder to distinguish \nc and \hbc \models. 

(3)~We show an important threat of this vulnerability in a recent approach where knowledge distillation~\cite{hinton2015distilling} is used to train a student \model on private unlabeled data via a teacher \model that is already trained on a set of labeled data, and show that an \hbc teacher can transfer its curiosity capability to the student \models. 

(4)~We support our findings via several experimental results on two real-world datasets with different characteristics, as well as additional analytical results for the setting of training convex \models. 

Code and instructions for reproducing the reported results are available at 
\textcolor{blue}{\mbox{\href{https://github.com/mmalekzadeh/honest-but-curious-nets}{https://github.com/mmalekzadeh/honest-but-curious-nets}.}}

\section{Problem Formulation}\label{sec:problem}
{\bf Notation.\label{notations}} We use lower-case {\it italic}, \eg $x$, for scalar variables; upper-case {\it italic}, \eg $X$, for scalar constants; lower-case bold, \eg $\bx$, for vectors; 
upper-case blackboard, \eg $\bbX$, for sets; calligraphic font, \eg $\mathcal{X}$, for functions; subscripts, \eg $w_1$, for indexing a vector; superscripts, \eg $w^1$, for distinguishing different instances; $\bbR^{X}$ for real-valued vectors of dimension $X$; and $\bbP^{X}$  for a probability simplex of dimension $X-1$ (that denotes the space of all probability distributions on a $X$-value random variable). We have $\{0,\ldots,X-1\}\equiv [X]$, and $\round{\cdot}$ shows rounding to the nearest integer. Logarithms are natural unless written explicitly otherwise. 
The standard logistic is $\sigma(\bx) = {1}/\big({1+\exp(-\bx)}\big)$.
Given random variables $a \in [X]$, $b \in [X]$, and $c \in [Z]$, the entropy of $a$ is $\tH(a) = -\sum_{i=0}^{X-1} \mathrm{Pr}(a=i) \log\mathrm{Pr}(a=i)$, the cross entropy of $b$ relative to $a$ is $\tH_{a}(b) = -\sum_{i=0}^{X-1} \mathrm{Pr}(a=i) \log\mathrm{Pr}(b=i)$, the conditional entropy of $a$ given $c$ is $\tH(a|c) = -\sum_{i=0}^{X-1}\sum_{j=0}^{Z-1} \mathrm{Pr}(a=i,c=j) \log\big(\mathrm{Pr}(a=i,c=j)/\mathrm{Pr}({c=j})\big)$, and the mutual information~(MI) between $a$ and $c$ is $\tI(a;c) = \tH(a) - \tH(a | c)$~\cite{mackay2003information}. $\bbI_{(\mathsf{C})}$ shows the indicator function that outputs $1$ if condition $\mathsf{C}$ holds, and $0$ otherwise.

{\bf Definitions.} Let a {\em \user} own data $\bx \in \bbR^{M}$ sampled from an unknown data distribution $\cD$. Let $\bx$ be informative about at least two latent categorical variables ({\em attributes}): $y \in [Y]$ as the {\em \pub} attribute, and $s \in [S]$ as the {\em \priv} attribute (see Figure~\textcolor{blue}{\ref{fig:classifier}}). Let a \server own a {\em \model} $\cF$ that takes $\bx$ and outputs: 
$   
\byh = \cF(\mathbf{\bx})  = [ \yh_0,  \yh_1, \ldots,  \yh_{Y-1}], 
$
where $ \yh_i$ {\em estimates} $Pr(y=i|\bx)$. Let $\yb$ denote the predicted value for $y$ that is decided from $\byh$; \eg based on a {\em threshold} in binary classification or {\em argmax} function in multi-class classification. We assume that $\bx$, and all intermediate computations of $\cF$, are hidden and the \user only releases $\byh$. Let $\bsh = \cG(\byh)$ be the {\em attack} on the \priv attribute $s$ that only the \server knows about. Let $\sb$ denote the predicted value for $s$ that is decided based on $\bsh$. In sum, the following Markov chain holds: $(y,s) \rightarrow \bx \rightarrow \byh \rightarrow \bsh$. 

Throughout this paper, we use the following terminology:  

{\bf 1. Honesty.} Given a test dataset $\bbD^{test} \sim \cD$, we define $\cF$ as a \dhon \model if
$$\Pr_{(\bx, y) \sim \bbD^{test},\text{ } \yb \leftarrow \cF(\bx)} [ \yb = y]  \geq \delta^y, $$ 
where $\delta^y \in [0,1]$ is known as the \model's  test accuracy, and we call it the {\em honesty} of $\cF$ in predicting the {\em \pub} attribute. 

{\bf 2. Curiosity.} Given a test dataset $\bbD^{test}\sim \cD$ and an  attack  $\cG$, we define $\cF$ as a \dcur \model if
$$\Pr_{(\bx, s) \sim \bbD^{test},\text{ } \sb \leftarrow \cG\big(\cF(\bx)\big)} [ \sb = s]  \geq \delta^s, $$  
where $\delta^s \in [0,1]$ is the attack's success rate on the test set, and we call it the {\em curiosity} of $\cF$ in predicting the {\em \priv} attribute.

{\bf 3. Honest-but-Curious~(\hbc).} We define $\cF$ as a $(\delta^y, \delta^s)$-\hbc \model if it is both \dhon and \dcur on the same $\bbD^{test}$. 

{\bf 4. {\bf \Nc} \Model.} A \model $\cF$ that is trained only for achieving the best honesty, without any intended curiosity.

{\bf 5. Black- vs. White-Box.} We consider the \users' perspective to the \model $\cF$ at {\em inference} time. In a {\em black-box} view, a \user observes only the \model's outputs and not the \model's architecture and parameters.
In a {\em white-box} view, a \user also has full access to the \model's architecture, parameters, and intermediate computations.

{\bf 6. Threat Model.}  The semi-trusted \server chooses the algorithm and dataset~($\bbD^{train}\sim \cD$) for training $\cF$. In a black-box view, the \server has the additional power to choose the architecture of $\cF$ (unlike the white-box view). At inference time, a \user (who does not necessarily participate in the training dataset) runs the trained \model on her private data once, and only reveals the \model's outputs $\byh=\cF(\bx)$ to the \server. We assume no other information is provided to the \server at inference time.

{\bf Our Objective.} We show how a \server can train an \hbc \model to establish efficient {\em honesty-curiosity} trade-offs over achievable $(\delta^y, \delta^s)$ pairs, and analyze the privacy risks, behavior, and characteristics of \hbc \models compared to \nc ones.  

\section{Black-box View: A Mixture Model}  \label{sec:black-box}
To build a better intuition, we first discuss the {\em black-box} view where the \server can choose any arbitrary architecture. and we show the existence of an efficient attack for every task with $S\leq Y$. 
\subsection{A Convex \Model}
We start with a simple logistic regression \model. Let us consider the synthetic data distribution depicted in Figure~\textcolor{blue}{\ref{fig:syn_data}}, where each sample $\bx \in \bbR^2$ has two attributes $y \in \{0,1\}$ and $s \in \{0,1\}$. If $s$ is correlated with the $y$, the output of any \model always reveals some \priv information (which we explore it in real-world datasets in Section~\textcolor{blue}{\ref{sec:eval}}) . The less $s$ is correlated with the $y$, the more difficult it should be for the \server to build an \hbc \model.  Thus, the data distribution in Figure~\textcolor{blue}{\ref{fig:syn_data}} is built such that $y$ and $s$ are independent, and for each attribute, there is an optimal linear classifier.    

It is clear that for this dataset, we can find an optimal logistic regression  \model $\yh = \sigma(w_1x_1 + w_2x_2+b)$, with parameters $[w_1, w_2, b]$, that simulates the decision boundary of $y$. Such a \model is \dhon with $\delta^y=1$, and at the same time it is \dcur with $\delta^s=0.5$; that means the \model is honest and does not leak any \priv information. The main point is that any effort for making a curious linear \model with $\delta^s>0.5$ will hurt the honesty by forcing  $\delta^y<1$. On this dataset, it can be shown that for any logistic regression \model we have $\delta^y=1.5-\delta^s$. For example, the optimal linear \model for attribute $s$ cannot have a better performance than a random guess on attribute $y$.

In Appendix~\textcolor{blue}{\ref{sec:apx:convex}}, we show how a logistic regression \model can become \hbc with a convex loss function, and analyze the behavior of such a \model in detail. Specifically, we show that the trade-off for a \model with {\em limited} capacity (\eg logistic regression) is that: if alongside the \pub attribute, we also optimize for the \priv attribute, we will only ever converge to a neighborhood of the optimum for the \pub attribute. We show that the size of the neighborhood is getting larger by the weight (\ie importance) we give to curiosity. While the analysis in Appendix~\textcolor{blue}{\ref{sec:apx:convex}} holds for a convex setting and are simplistic in nature, it provides intuitions into the idea that when the attributes $y$ and $s$ are somehow {\em correlated}, the output $\yh$ can better encode both tasks, but when we have {\em independent} attributes, we need \models with more capacity to cover the payoff for not converging to the optimal point of \pub attribute.

\subsection{A Mixture of Two \Models}
\begin{figure}[t]
    \centering
    \includegraphics[width=.65\columnwidth]{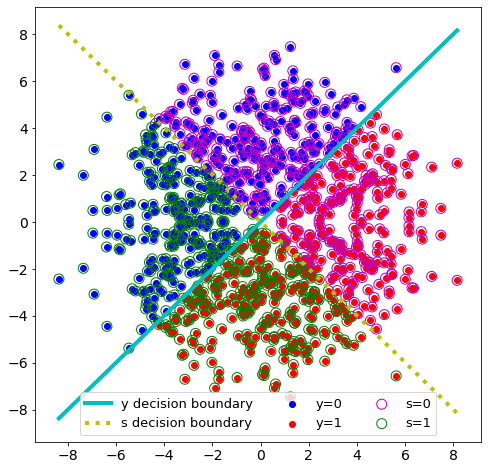}
    \vspace{-.3cm}
    \caption{A two-attribute two-class data distribution on $\bbR^2$, where attributes $y$ and $s$ are independent, and classes in each attribute are linearly separable. \label{fig:syn_data}} 
    \vspace{-.3cm} 
\end{figure}

\begin{figure}[t] 
    \includegraphics[width=.6\columnwidth]{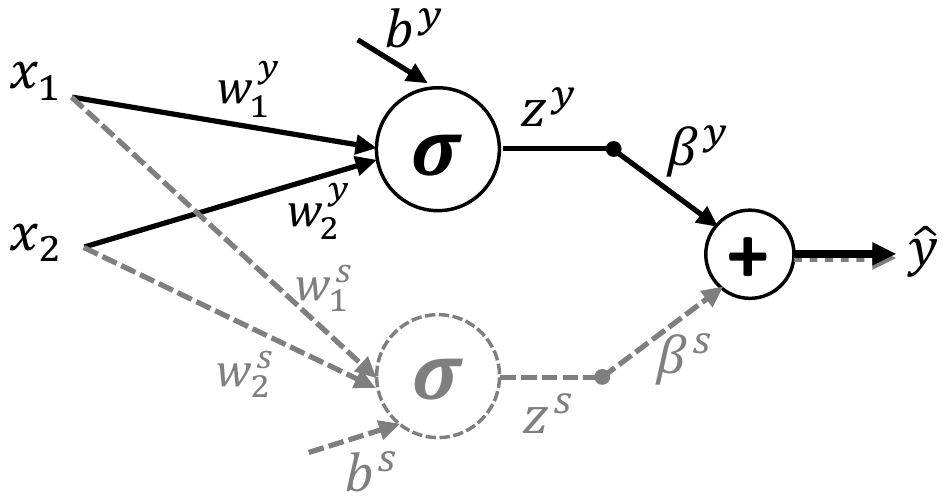}
    \vspace{-.3cm}
    \caption{Two logistic regression \models, each trained on $\bx$ separately: $z^{y}$ for $y$ and $z^{s}$ for $s$. \Models outputs are mixed with each other such that the final output, $\yh$, is a real-valued scalar informative about both attributes. 
     \label{fig:basic_model}}
    \begin{center}
    \begin{minipage}[t]{.7\columnwidth}
    \includegraphics[width=\columnwidth]{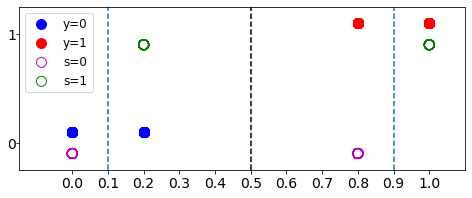}
    \end{minipage}
    \vfill
    \begin{minipage}[t]{.7\columnwidth}
    \includegraphics[width=\columnwidth]{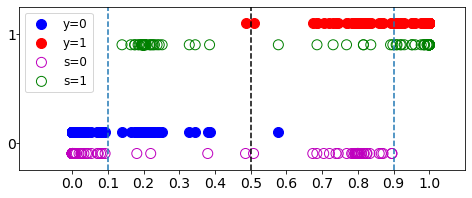}
    \end{minipage}
    \vspace{-.3cm}
    \caption{ Produced $\yh$ in Figure~\textcolor{blue}{\ref{fig:basic_model}} for the dataset in Figure~\textcolor{blue}{\ref{fig:syn_data}}, using $\beta^y = 0.8$ and $\beta^s = 0.2$. (Top) the hard mixture achieves $\delta^y=1$ and $\delta^s=1$ (Eq~\eqref{eq:hard_mix}). (Bottom) the normal mixture achieves $\delta^y=.99$ and $\delta^s=.95$ using $\tau'=.1$ (Eq~\eqref{eq:tau_binary}). \label{fig:basic_model_data_preds}}
    \vspace{-.5cm} 
    \end{center}
\end{figure}

Figure~\textcolor{blue}{\ref{fig:basic_model}} shows how two logistic regression \models, each trained separately for a corresponding attribute, can be combined such that the final output $\yh$ is a mixture of the predicted values for the \pub attribute $y$, and the \priv attribute $s$. There are two ways to combine $z^y \in [0,1]$ and $z^s\in [0,1]$. Considering multipliers $\beta^y \in [0,1]$ and $\beta^s = 1-\beta^y$, one option is the {\em normal} mixture, where $\yh = \beta^{y}z^{y} + \beta^{s}z^{s}$, and another one is the {\em hard} mixture, where $\yh = \beta^{y}\round{z^{y}} + \beta^{s}\round{z^{s}}$. Since two \models are each optimal for the dataset in Figure~\textcolor{blue}{\ref{fig:syn_data}}, $\yh$ in hard mixture can only take four values: 
\begin{equation}\label{eq:hard_mix}
\yh = \begin{cases}
    0  & \text{if}\enspace  y = 0 \enspace\&\enspace s = 0 \\
    \beta^s  & \text{if}\enspace y = 0 \enspace\&\enspace s = 1  \\
    \beta^y  & \text{if}\enspace y = 1 \enspace\&\enspace s = 0  \\
    1  & \text{if}\enspace y = 1 \enspace\&\enspace s = 1.
    \end{cases}
\end{equation} 
By choosing $\beta^y \neq 0.5$, given a $\hat y$, we can accurately estimate both $y$ and $s$, which results in a $(\delta^y=1, \delta^s=1)$-\hbc \model. Notice that the \model in Figure~\textcolor{blue}{\ref{fig:basic_model}} is not a linear \model anymore, but its capacity is just twice the capacity of the logistic regression. Thus, while keeping the same honesty $\delta^y=1$, we could improve curiosity from $\delta^s=0.5$ to $\delta^s=1$ just by doubling the \model's capacity. 

The normal mixture is challenging as the range of possible values for $\yh$ is $[0,1]$. An idea is to define a threshold  $\tau'\in [0,1]$ and divide the range of $[0,1]$ into four sub-ranges such that:
\begin{equation}\label{eq:tau_binary}
\text{if } \begin{cases}
    \yh \in [0, \tau')  & \text{then we predict }  \yb = 0 \enspace\&\enspace \sb = 0 \\
    \yh \in [\tau', 0.5)  & \text{then we predict }  \yb = 0 \enspace\&\enspace \sb = 1 \\
    \yh \in [0.5, 1-\tau')  & \text{then we predict }  \yb = 1 \enspace\&\enspace \sb = 0\\
    \yh \in [1-\tau', 1)  & \text{then we predict }  \yb = 1 \enspace\&\enspace \sb = 1.
    \end{cases}
\end{equation} 
As we see in Figure~\textcolor{blue}{\ref{fig:basic_model_data_preds}}, a normal mixture cannot guarantee the optimal $(\delta^y=1, \delta^s=1)$-\hbc that we could obtain via a hard mixture.
Nevertheless, in the following sections, we will show that the idea of dividing  the range $[0,1]$ into four sub-ranges is still useful, especially in white-box situations, where a hard mixture approach is not an option but we can have non-linear \models.  

{\bf Summary.} In a black-box view the \server can always train two separate \models, each with sufficiently high accuracy, and can use the hard mixture of two outputs to build the best achievable $(\delta^y, \delta^s)$-\hbc \model; no matter what type of \model is used.  Basically, the \server can mostly get the same performance if it could separately run two \models on the data. We emphasize that the \server's motivation for such a mixture \model (and not just simply using two separate \models) is that the shape of the \model's output $\byh$ depends on $Y$. Thus, a limitation is that such an attack works only if $S \leq Y$, which in general is not always the case. For instance, let $Y=5$, $S=3$, $\beta^y = 0.8$, and $\beta^s=0.2$. If the hard outputs for $y$ is $\round{\bz^y} = [0,0,0,1,0]$ and for $s$ is $\bz^s = [0,1,0]$, then by observing ${\byh}=[0,0.2,0,0.8,0]$, \server can estimate both $y$ and $s$ while looking very honest. But if $S>Y$, then the \server cannot easily encode the private attribute via a mixture model, because we assume that the cardinality of the \model's output is limited to $Y$ as it is supposed to look like a \nc \model. Thus, situations with $S > Y$ are more challenging, particularly when \users have a white-box view and  the \server is not free to choose any arbitrary architecture, \eg it has to train an of-the-shelf \model. In the following, we focus on the  white-box view in a general setting.

\section{White-box View: A General Solution}\label{sec:white-box}
Theoretically, the \model's output $\byh$, as a real-valued vector, can carry an infinite amount of information. Thus, releasing $\byh$ without imposing any particular constraint can reveal any private information and even can be used to (approximately) reconstruct data $\bx$. One can imagine a hash function that maps each $\bx$ to a specific $\byh$, and consequently, by observing $\byh$, we can reconstruct $\bx$~\cite{radhakrishnan2020overparameterized}. However, the complexity of real-world data, assumptions on the required honesty, white-box view, requirement of \Pval outputs, and other practical constraints will rule out such trivial solutions. Here, we discuss the connection between the curiosity of a \model and the entropy of its output, and then we formulate the problem of establishing a desired trade-off for an \hbc \model $\cF$ and its corresponding attack $\cG$ into an information-theoretical framework.

\subsection{Curiosity and Entropy}
The entropy of a random variable $x$ is the expected value of the information content of that variable; also called self-information: $\tH(x) = \tI(x;x)$.
When we are looking for \pub information in the data, then the presence of other potentially unrelated information in that data could make the extraction of \pub information more challenging. In principle, such unrelated information would act as noise for our target task. For example, when looking for a \pub attribute, a trained DNN \model takes data $\bx$ (usually with very high entropy) and produces a probability distribution over the possible outcomes $\byh \in [Y]$ with much lower entropy compared to $\bx$. Although $\byh$ contains much less information than $\bx$ in the sense that it has less entropy, it is considered more informative \wrt the target $y$. 

On the other hand, there is a relationship between the entropy of a \model's output,  $\tH(\byh)\in [0,\log(Y)]$, and the curiosity  $\delta^s$ of an  attack  $\cG$. 
The larger $\tH(\byh)$, the more information is carried by $\byh$, thus the higher the chance to reveal information unrelated to the \pub task. For example, assume that $Y=4$, $y=1$, and $y$ is independent of $s$. In the extreme case when the \model's output is $\byh=[0,1,0,0]$ (that means $\tH(\byh)=0$), then $\byh$ carries no information about $s$ and adding any information about $s$ would require increasing the entropy of the output $\byh$. 

In supervised learning, the common loss function for training DNN \models is {\em cross entropy}: $\tH_y(\byh) = -\sum_{i=0}^{Y-1} \bbI_{(y=i)}\log \yh_i$; that inherently minimizes $\tH(\byh)$ during  training. However, since data is usually noisy, we cannot put any upper bound on $\tH_y(\byh)$ at inference time. In practice, minimizing $\tH(\byh)$, alongside $\tH_y(\byh)$, might help in keeping $\tH(\byh)$ low at inference time, which turns out to be useful for some applications like semi-supervised learning~\cite{grandvalet2005semi}. But there is no guarantee that a \model will always produce a minimum- or bounded-entropy output at inference time. This fact somehow serves as the main motivation of our work for encoding private attributes of the classifier's input into the classifier's output; explained in the following two attacks. 

\subsection{Regularized Attack}\label{sec:sol_bin}

We first introduce a method, for training any \model to be \hbc, in situations where \priv attribute is binary ($S=2$) and the \server only has access to the \Pval output ($\byh \in \bbP^{Y}$); see Figure~\textcolor{blue}{\ref{fig:classifier}}.  The idea is to enforce \model $\cF$ to explicitly encode $s$ into the entropy of $\byh$ by {\em regularizing} the loss function on $\cF$. In general, there are two properties of $\byh$ that one can utilize for creating an \hbc \model:

{\bf 1. Argmax}: as the usual practice, we use the index of the maximum element in $\byh$ to predict $y$. This helps the \model to satisfy the honesty requirement. 

{\bf 2. Entropy}: the entropy of $\byh$ can have at least two states: (i) be close to the maximum entropy, \ie $\tH(\byh)=\log Y$, or (ii) be close to the minimum entropy, \ie $\tH(\byh)=0$.

\begin{figure}[t]
    \centering
    \includegraphics[width=.6\columnwidth]{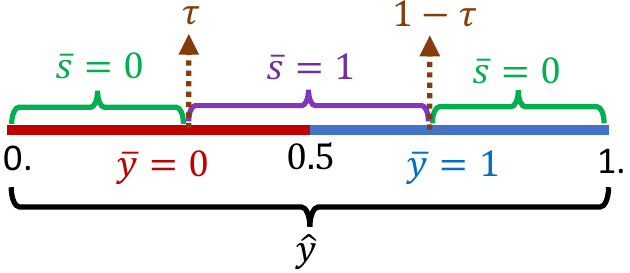}
    \vspace{-.3cm}
    \caption{Observing $\yh \in [0,1]$, we can predict $\yb=0$ if $\yh < .5$, otherwise $\yb=1$. We train the \model such that for samples with $s=0$, $\yh$ gets close to the borders (0 or 1) depending on $y$; otherwise, for $s=1$, $\yh$ gets far from the borders. Using a threshold $\tau$, we predict binary attributes $y$ and $s$ from $\yh$.}
    \label{fig:binary} 
    \vspace{-.4cm} 
\end{figure}

\noindent We show that $\tH(\byh)$ can be used for predicting a binary $s$, while not interfering with the $\mathtt{argmax}(\byh)$ that is preserved for $y$. Without loss of generality, let us assume $Y=2$. Consider observing $\yh \in \bbP$; that is equivalent to $\byh =[\yh_0,\yh_1] \in \bbP^{2}$, when $\yh_1\equiv\yh$ and $\yh_0=1-\yh_1$. Figure~\textcolor{blue}{\ref{fig:binary}} shows how we can use the single real-valued $\yh$ to predict two attributes. For example, $\byh=[.95, .05]$ and $\byh=[.75, .25]$ have the same $\mathtt{argmax}$ but different entropies: $0.29$ and $0.81$, respectively.
    
{\bf Training.} Choosing any arbitrary classifier $\cF$, the \server can train $\cF$ with the following loss function:
\begin{equation}\label{eq:loss_func_binary}
\begin{split}
   & \cL^{b} =  \beta^{y} \tH_{y}(\byh) + \beta^{s}(\bbI_{(s = 0)}
    - \bbI_{(s = 1)})\tH(\byh) = 
    \\
    &  
    \beta^{y}\big(-\sum_{i=0}^{Y-1}  \bbI_{(y=i)}\log{ \yh_i}\big)
    + \beta^{s}\big(\bbI_{(s = 0)}
    - \bbI_{(s = 1)}\big)\big(-\sum_{i=0}^{Y-1}  \yh_i\log{ \yh_i}\big), 
\end{split}
\end{equation}
where multipliers $\beta^{y}$ and $\beta^{s}$ aim to control the trade-off between honesty and curiosity. In Eq~\eqref{eq:loss_func_binary}, in the first term, we have the cross-entropy and in the second term, we have Shannon entropy that aims to minimize the entropy of $\byh$ for samples of $s = 0$, while maximizing the entropy of $\byh$ for samples of $s = 1$.  

{\bf Attack.} At inference time, when the \server observes $\byh$, it computes
$\tH(\byh )$, and using a threshold $\tau \in [0,1]$, estimates $\sb$:
\begin{equation}\label{eq:infer_s_bin}
 \sb= \cG(\byh) =\left\{
  \begin{array}{@{}ll@{}}
    0, & \text{if}\ \tH(\byh) \leq \tau  \\
    1, & \text{otherwise.}
  \end{array}\right.   
\end{equation}
Thus, the attack $\cG$ is a simple threshold function, and $\tau$ is optimized using the validation set during training, as we explain in Section~\textcolor{blue}{\ref{sec:eval}}.

\subsection{Parameterized Attack}\label{sec:general_cat}
In this section, we present our general solution that works for $S\geq2$, and for both \Rval and \Pval outputs; see Figure~\textcolor{blue}{\ref{fig:classifier}}. 
 
\subsubsection{An Information Bottleneck Formulation}\label{sec:IB}
Remember the Markov chain $(y,s) \rightarrow \bx \rightarrow \byh \rightarrow \bsh$. We assume that the \server is constrained to a specific family of \models~$\bbF$, \eg a specific DNN architecture that has to be as honest as a \nc \model. The \server looks for a $\cF^* \in \bbF$ that maps the \users' data $\bx$ into a vector $\byh$ such that $\byh$ is as informative about both $y$ and $s$ as possible. 

Formally,  $\cF^*$ can be defined as the solution of the following mathematical optimization:
\begin{equation}\label{eq:optim_F_1} 
\min_{(\bx, y,s) \leftarrow \cD, \text{ } \cF \in \bbF, \text{ } \byh \leftarrow \cF(\bx)} \big[\cI = \beta^{x}\tI(\byh; \bx) -\beta^{y}\tI(\byh; y) - \beta^{s}\tI(\byh;s)\big],
\end{equation}
where $\beta^{x} $, $\beta^{y}$, and $\beta^{s}$ are Lagrange multipliers that allow us to move along different possible local minimas and all are non-negative real-valued\footnote{Mathematically speaking, we only need two Lagrange multipliers as $\beta^y$ and $\beta^s$ are dependent. Here we use a redundant multiplier for the ease of presentation.}.  Eq~\eqref{eq:optim_F_1} is an extension of the {\em information bottleneck}~(IB)  formulation~\cite{tishby2000information}, where the optimal $\byh$, produced by $\cF^*$, is decided based on its relation to the three variables, $\bx$, $y$, and~$s$.  By varying the $\beta$ multipliers, we can explore
the trade-off between {\em compression} at various rates, \ie by minimizing $\tI(\byh; \bx)$, and the amount of information we aim to preserve, \ie by maximizing $\tI(\byh; y)$ and $\tI(\byh;s)$. Particularly for DNNs, it is shown that compression might help the \model to achieve better generalization~\cite{tishby2015deep}.

{\bf An Intuition.} For better understanding, assume that for $\beta^s=0$ (\ie the \nc \models) the optimal solution $\cF^*$ obtains a specific value for $\cI=\cI^*$ in Eq~\eqref{eq:optim_F_1}. Now, assume that the \server aims to find an \hbc solution, by setting $\beta^s>0$. Then, in order to maintain the same value of $\cI^*$ with the same MI for the \pub attribute $\tI(\byh; y)$, the term $\tI(\byh; \bx)$ must be increased, because $\tI(\byh;s) \geq 0$ if $\beta^s>0$. Since for deterministic \models $\tI(\byh; \bx)=\tH(\byh)$, we conclude that when the \server wants to encode information about both $y$ and $s$ in the output $\byh$, then the output's entropy must be higher than if it was only encoding information about $y$; which shows that our formulation in Eq~\eqref{eq:optim_F_1} is consistent with our motivation for exploiting the capacity of $\tH(\byh)$. In Section~\textcolor{blue}{\ref{sec:eval}}, we provide more intuition on this through some experimental results.

\subsubsection{Variational Estimation}\label{sec:variational} 
Now, we analyze a \server that is computationally bounded and has only access to a sample of the true population (\ie a training dataset $\bbD^{train}$) and wants to solve Eq~\eqref{eq:optim_F_1} to create a (near-optimal) \hbc \model $\cF$. We have
\begin{equation*}
\begin{split}
& \cI = \beta^{x}\tI(\byh, \bx)-\beta^{y}\tI(\byh; y) - \beta^{s}\tI(\byh;s) =
  \\ 
& \beta^{x}\tH(\byh) -\beta^{x}\tH(\byh|\bx)-\beta^{y}\tH(y)+\beta^{y}\tH(y|\byh)-\beta^{s}\tH(s)+\beta^{s}\tH(s|\byh).
\end{split}
\end{equation*}
Since for a fixed training dataset $\tH(y)$ and $\tH(s)$ are constant during the optimization and for a deterministic $\cF$ we have $\tH(\byh|\bx)=0$, we can simplify Eq~\eqref{eq:optim_F_1} as
\begin{equation}\label{eq:optim_F_2} 
\min_{(\bx, y,s) \leftarrow \cD, \text{ } \cF \in \bbF, \text{ } \byh \leftarrow \cF(\bx)} \big[\cH = \beta^{x}\tH(\byh) + \beta^{y}\tH(y|\byh) +  \beta^{s}\tH(s|\byh)\big].  
\end{equation} 
Eq~\eqref{eq:optim_F_2} can be interpreted as an optimization problem that aims to minimize the entropy of $\byh$ subject to encoding as much information as possible about $y$ and $s$ into $\byh$.
Thus, the optimization seeks for a function $\cF^*$ to produce a low-entropy $\byh$ such that $\byh$ is only informative about $y$ and $s$ and no information about anything else. Multipliers $\beta^{y}$ and $\beta^{s}$ specify how $y$ and $s$ can compete with each other for the remaining capacity in the entropy of $\byh$; that is challenging, particularly, when $y$ and $s$ are independent.

Different constraints on the server can lead to different optimal models. As we observed, in a black-box view with $S \leq Y$ and arbitrary $\bbF$, a solution is achieved by training two separate classifiers with a cross-entropy loss function and an entropy minimization regularizer~\cite{grandvalet2005semi}. First, using stochastic gradient decent~(SGD), we train \model $\cF^{y}$ by setting $\beta^{y}=1$ and $\beta^{s}=0$ in Eq~\eqref{eq:optim_F_2}. Second, we train \model $\cF^{s}$ by setting $\beta^{y}=0$ and $\beta^{s}=1$. Finally, we build $\cF=\beta^{y}\round{\cF^{y}}+\beta^{s}\round{\cF^{s}}$ as the desired \hbc \model for any choice of $\beta^{y}\in[0,1]$ and $\beta^{s}=1-\beta^{y}$. Notice that the desired value for $\beta^x$ can be chosen through a cross-validation process. 

Thus, let us focus on the white-box view with a constrained $\bbF$, \eg where the \server is required to train a known off-the-shelf \model. Here, we use the cross-entropy loss function for the \pub attribute $y$ and train the  \model $\cF$ using SGD. However, besides this cross-entropy loss, we also need to look for another loss function for the attribute $s$. Thus, we need a method to simulate such a loss function for $s$. 

Let $p_{s | \byh}$ denote the true, but unknown, probability distribution of $s$ given $\byh$, and $q_{s | \byh}$ denote an approximation of $p_{s | \byh}$. Considering the cross entropy between these two distributions $\tH_{p_{s | \byh}}(q_{s | \byh})$, it is known~\cite{mcallester2020formal} that
\begin{equation}\label{eq:ce_upb}
\tH(p_{s | \byh})=
-\sum_{i=0}^{S-1} p_{s | \byh} \log(p_{s | \byh}) 
\leq  
\tH_{p_{s | \byh}}(q_{s | \byh})=
-\sum_{i=0}^{S-1} p_{s | \byh} \log(q_{s | \byh}).     
\end{equation}
This inequality tell us that the cross entropy between the unknown distribution, \ie $p_{s | \byh}$, and any estimation of it, \ie $q_{s | \byh}$, is an upper-bound on $\tH(p_{s | \byh})$; and the equality holds when $q_{s | \byh} = p_{s | \byh}$. Thus, if we find a useful model for $q_{s | \byh}$, then the problem of minimizing $\tH(p_{s | \byh})$ can be solved through minimization of $\tH_{p_{s | \byh}}(q_{s | \byh})$. 
 
\begin{figure}[t]
    \centering
    \includegraphics[width=.85\columnwidth]{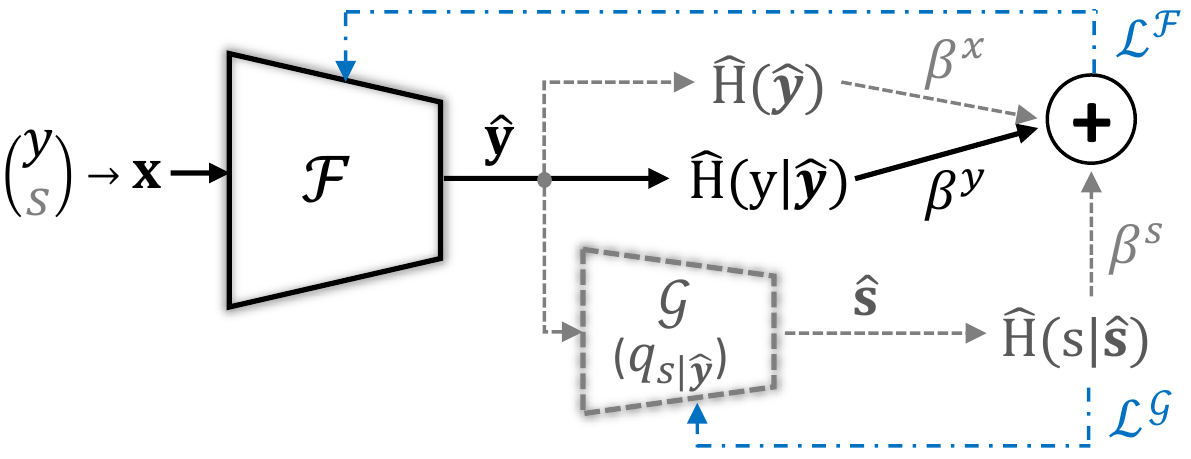}
    \caption{The \server iteratively trains the chosen \model $\cF$ and attack $\cG$ that is a {\em parameterized} model. While $\cG$ is optimized only based on $\hat{\tH}(s|\sh)$, \model $\cF$ is optimized based on a weighted combination of $\hat{\tH}(\byh)$, $\hat{\tH}(y|\byh)$ and $\hat{\tH}(s|\byh)$ in  Eq~\eqref{eq:optim_F_2_emp}.}  
    \label{fig:adv_train} 
\end{figure}

{\bf Training.} In practice, parameterized models such as neural networks, are suitable candidates for $q_{s | \byh}$~\cite{poole2019variational, mcallester2020formal}. For $N$ i.i.d. samples of pairs $\big\{(\byh^1, s^1)$, $\ldots$, $(\byh^N, s^N)\big\}$ that represent $p_{s | \byh}$ and are generated via our current \model $\cF$ on a dataset $\bbD^{train}\sim\cD$, we can estimate $\tH_{p_{s | \byh}}(q_{s | \byh})$ using the empirical cross-entropy as
\begin{equation}\label{eq:emp_cross_ent} 
    \hat{\tH}^{N}_{p_{s | \byh}}(q_{s | \byh})= -\frac{1}{N} \sum_{n=1}^{N} \sum_{i=0}^{S-1} \bbI_{(s=i)}\log\big( q_{s|\byh}(\byh^n)\big).
\end{equation}  
Therefore, after initializing a {\em parameterized} model for $q_{s | \byh}$ to estimate $\tH(p_{s | \byh})$, we run  optimization
\begin{equation}\label{eq:minimize_cond_ent}
    \min_{q_{s|\byh}} \hat{\tH}^{N}_{p_{s | \byh}}(q_{s | \byh}) ,
\end{equation}
where we iteratively sample $N$ pairs $(s^i,\byh^i)$, compute Eq~\eqref{eq:emp_cross_ent}, and update the model $q_{s|\byh}$; using SGD. The \server will use this additional model $q_{s|\byh}$ for $s$, alongside the cross entropy loss function for $y$, to solve Eq~\eqref{eq:optim_F_2} for finding the optimal $\cF$. Considering $q_{s|\byh}$ as the attack $\cG$ and setting our desired $\beta$ multipliers, the variational approximation of our general optimization problem in Eq~\eqref{eq:optim_F_2} is written as
\begin{equation}\label{eq:optim_F_2_emp} 
\min_{(\bx, y,s) \leftarrow \cD, \text{ } \cF \in \bbF, \text{ } \byh \leftarrow \cF(\bx), \bsh = \cG(\byh)} \big[\cH = \beta^{x}\hat{\tH}(\byh) + \beta^{y}\hat{\tH}(y|\byh) +  \beta^{s}\hat{\tH}(s|\byh)\big],
\end{equation} 
where the joint minimization is performed over both parameterized models $\cF$ and $\cG$. Here, $\hat{\tH}(\cdot)$ and $\hat{\tH}(\cdot|\cdot)$ denote the empirical entropy and conditional-entropy computed on every sampled batch of data, respectively. A schematic view of this approach is depicted in Figure~\textcolor{blue}{\ref{fig:adv_train}}.  Using a training dataset $(\bx,y,s)\in\bbD^{train}\sim\cD$ and in an iterative process, both \model $\cF$, with loss function $\cL^{\cF} = \beta^{x}\hat{\tH}(\byh) + \beta^{y}\hat{\tH}(y|\byh) +  \beta^{s}\hat{\tH}(s|\byh),$ and  attack  $\cG$, with loss function $\cL^{\cG} = \hat{\tH}(s|\byh)$, are simultaneously trained. 
Algorithm~\ref{alg:fg_ib}, in Appendix~\textcolor{blue}{\ref{appx:algo}}, shows the details of our proposed training method.  

{\bf Summary.} While our {\em regularized} attack~(Section~\textcolor{blue}{\ref{sec:sol_bin}}) works by only modifying the loss function of $\cF$, in our {\em parameterized} attack (this section) we not only need to modify the loss function of $\cF$ but also to utilize an additional model $\cG$ (\eg a multi-layer perception) to estimate the sensitive attribute. Hence, $\cF$ can be seen as a model that aims to perform two roles at the \user's side: (1) as an honest {\em classifier}, to estimate the \pub attribute, and (2) as a curious {\em encoder} to encode the \priv attribute. At the \server side, $\cG$ acts as a {\em decoder} for decoding the received output for estimating the \priv attribute. Notice that, since the sensitive attribute is a categorical variable, we cannot directly reconstruct $s$, \eg using mean squared error in traditional autoencoders, and we need Equation~\eqref{eq:minimize_cond_ent}.

\section{Evaluation}\label{sec:eval} 
We present results in several tasks from two real-world datasets and discuss the honesty-curiosity trade-off for different attributes.

\subsection{Experimental Setup} 
\subsubsection{Settings} We define a {\em task} as training a specific \model $\cF$ on a training dataset including samples with two attributes, $y\in[Y]$ and $s\in[S]$, and evaluating $\cF$ on a test dataset by measuring honesty $\delta^y$ and curiosity $\delta^s$ via the attack $\cG$ (see Section~\textcolor{blue}{\ref{sec:problem}}). We have two types of attacks: (i) {\em regularized}, as explained in Section~\textcolor{blue}{\ref{sec:sol_bin}}, and (ii) {\em parameterized}, as explained in Section~\textcolor{blue}{\ref{sec:general_cat}}. For each task there are 3 different scenarios: (1) \NC: where $\cF$ is a standard \model without intended curiosity, (2) \HBCR: where $\cF$ is trained to be \hbc and $\cG$ has access to \Rval outputs $\byh\in \bbR^Y$, and (3) \HBCP: where $\cF$ is \hbc but $\cG$ has access only to \Pval outputs $\byh\in \bbP^Y$ (see Figure~\textcolor{blue}{\ref{fig:classifier}}). We run each experiment five times, and report mean and standard deviation. For each experiment in \HBCP, $\tau\in[0,1]$ in Eq~\eqref{eq:infer_s_bin} is chosen based on the validation set and is used to evaluate the result on the test set. 

\subsubsection{CelebA Dataset~\cite{liu2015celeba}} This is a dataset including more than 200K celebrity face images, each with 40 binary attributes, \eg the `Smiling' attribute with values $0$:$Non\text{-}Smile$ or $1$:$Smile$. We choose attributes that are almost balanced, meaning that there are at least 30\% and at most 70\% samples for that attribute with value $1$. Our chosen attributes are: Attractive, BlackHair, BlondHair, BrownHair, HeavyMakeup, Male, MouthOpen, Smiling, and WavyHair. 
CelebA is already split into separate training, validation, and test sets. We use the resampled images of size $64\times64$. We elaborate more on CelebA and show some samples of this dataset in Appendix~\textcolor{blue}{\ref{apx:celeba}}.  

\subsubsection{UTKFace Dataset~\cite{zhang2017age}} This is a dataset including 23,705 face images annotated with attributes of Gender (Male or Female), Race (White, Black, Asian, Indian, or others), and Age (0-116). We use the resampled images of size $64\times64$, and randomly split UTKFace into subsets of sizes 18964 (80\%) and 4741 (20\%) for training and test sets, respectively.  A subset of 1896 images (10\%) from the training set is randomly chosen as the validation set for training. See Appendix~\textcolor{blue}{\ref{apx:utk}} for the details of tasks we define on UTKFace, and some samples.

\begin{table*}[t]
\caption{ (A) The characteristics of four attributes vs. Smiling attribute in the training set of CelebA dataset. The honesty  ($\delta^y$) and curiosity  ($\delta^s$) of (B) \nc \models when releasing either \Rval or \Pval outputs, as well as our proposed \hbc \models trained for (C) regularized attacks and (D) parameterized attacks. Here, $Y=S=2$ and $\beta^x= 0$. Values are in percentage (\%).} 
\label{tab:smile_vs_others}
\resizebox{\textwidth}{!}{
\begin{tabular}{@{}cccc|cccc|cccc|cccc@{}}
\multicolumn{16}{c}{{\bf(A)} For each attribute: the empirical joint probability and mutual information~(MI) with Smiling, besides accuracy of the \nc \model~($\delta^y$)}\\
\multicolumn{4}{c|}{\textbf{MouthOpen}} & \multicolumn{4}{c|}{\textbf{Male}} & \multicolumn{4}{c|}{\textbf{HeavyMakeup}} & \multicolumn{4}{c}{\textbf{WavyHair}}\\\hline
 & {Non-Smile} & {Smile} &  sum & & {Non-Smile} & {Smile} &   sum  & & {Non-Smile} & {Smile} &   sum & & {Non-Smile} & {Smile} &  sum \\ 
{0} & .386 & .122 & .508 &{0} & .292 & .375 & .667 & {0} & .305 & .184 & .489 & {0} & .322 & .166 & .488 \\ 
{1} & .103 &  .389 & .492 &  {1} & .197 & .136 & .333 & {1} & .226 & .28.5  & .511 & {1} & .303 & .209  & .512  \\
 sum &  .489 & .510 & &  sum & .489 & .511 & &  sum &  .531 & .469  & &  sum & .625 & .375 & \\\hline
& \multicolumn{3}{c|}{MI: $.231$  $\quad$ $\delta^y$: $93.42\pm.07$} &  \multicolumn{4}{c|}{MI: $.015$  $\quad$ $\delta^y$: $97.67\pm.04$} &  \multicolumn{4}{c|}{MI: $.024$ $\quad$ $\delta^y$: $88.89\pm.19$} & \multicolumn{3}{c}{MI: $.003$  $\quad$ $\delta^y$: $77.29\pm.55$} & \\
\bottomrule[1pt]  
\end{tabular} 
} 
\resizebox{\textwidth}{!}{
\begin{tabular}{@{}c|rccc|cc|cc|cc@{}}
 \multicolumn{1}{c}{}& \multicolumn{1}{c}{Setting} &$(\beta^y,\beta^s)$ & \multicolumn{2}{c}{$s$: \textbf{MouthOpen}} & \multicolumn{2}{c}{$s$: \textbf{Male}} &  \multicolumn{2}{c}{$s$: \textbf{HeavyMakeup}} &  \multicolumn{2}{c}{$s$: \textbf{WavyHair}}  \\\cline{2-11}
 \multicolumn{1}{c}{}&&&$\delta^y$&\multicolumn{1}{c}{$\delta^s$}&$\delta^y$&\multicolumn{1}{c}{$\delta^s$}&$\delta^y$&\multicolumn{1}{c}{$\delta^s$}&$\delta^y$&\multicolumn{1}{c}{$\delta^s$}\\
  \multicolumn{1}{c}{} & \multicolumn{10}{c}{{\bf(B)} Overlearning~\cite{song2019overlearning}}\\
\parbox[t]{2mm}{\multirow{10}{*}{\rotatebox[origin=c]{90}{$y$: \textbf{Smiling}}}} 
&
  \multirow{2}{*}{\NC}\Rval &  \multirow{2}{*}{(1., \textit{0.})} & \multirow{2}{*}{$92.15\pm.04$} & $\mathit{79.56\pm.32}$ & \multirow{2}{*}{$92.15\pm.04$} & $\mathit{68.20\pm.03}$  & \multirow{2}{*}{$92.15\pm.04$} & $\mathit{60.96\pm.10}$ & \multirow{2}{*}{$92.15\pm.04$} & $\mathit{57.93\pm.21}$ \\   
 & \Pval &  &  & $\mathit{79.22\pm.14}$ &  & $\mathit{68.20\pm.00}$  &  & $\mathit{60.53\pm.11}$ &  & $\mathit{57.90\pm.00}$ \\\cline{3-11}
 & \multicolumn{10}{c}{{\bf(C)} Regularized Attack}\\
 & \multirow{3}{*}{\HBCP} & (.7,.3) & $91.90\pm.01$ & $84.73\pm.50$     & $91.78\pm.15$ & $90.14\pm.37$     & $92.03\pm.12$ & $80.79\pm.47$ & $92.11\pm.15$ & $68.06\pm.72$ \\
 & & (.5, .5) & $91.83\pm.13$ & $89.08\pm.04$    & $91.65\pm.10$ & $94.20\pm.18$     & $91.87\pm.03$ & $85.27\pm.15$ & $91.98\pm.18$ & $72.36\pm.83$ \\  
 & & (.3, .7) & $91.75\pm.10$ &  $91.22\pm.25$    & $91.60\pm.11$ & $96.02\pm.09$     & $91.58\pm.10$ & $87.09\pm.43$  & $91.73\pm.07$ & $73.52\pm.77$ \\ \cline{3-11} 
& \multicolumn{10}{c}{{\bf(D)} Parameterized Attack}\\
&
 \multirow{1}{*}{\HBCR} & (.7,.3) & $91.84\pm.07$ & $93.40\pm.09$ & $92.36\pm.02$ &  $97.21\pm.10$ & $92.12\pm.09$ & $88.63\pm.04$ & $92.17\pm.04$ & $76.74\pm.37$ \\\cline{3-11}
  &  \multirow{3}{*}{\HBCP} & (.7,.3) &  $91.49\pm.44$ & $85.94\pm.76$   & $91.72\pm.17$ & $79.39\pm2.4$  & $91.84\pm.10$ & $84.45\pm.15$ & $92.12\pm.09$ & $57.90\pm.00$  \\
   & & (.5,.5) & $90.20\pm1.1$ & $88.91\pm.83$ & $90.17\pm.66$ & $92.73\pm.76$  & $91.49\pm.11$ & $86.19\pm.18$ & $92.20\pm.12$ & $61.51\pm.77$ \\
   & & (.3,.7) & $82.55\pm.27$ & $93.15\pm.07$ & $67.65\pm6.4$ & $97.41\pm.06$  & $70.75\pm4.3$ & $88.62\pm.52$ & $89.27\pm1.9$ & $67.57\pm3.7$ \\\cline{2-11} 
\end{tabular}  
} 
\end{table*}

\begin{figure}[t]
    \begin{center}
    \begin{minipage}[t]{.45\columnwidth}
    \centering
    \includegraphics[width=\columnwidth]{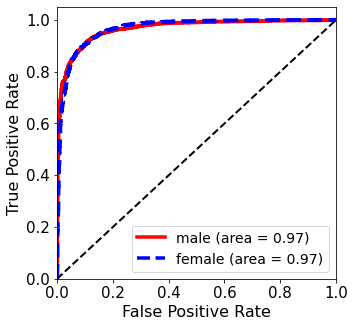}
    \end{minipage}
    \hfill
    \begin{minipage}[t]{.45\columnwidth}
    \centering
    \includegraphics[width=\columnwidth]{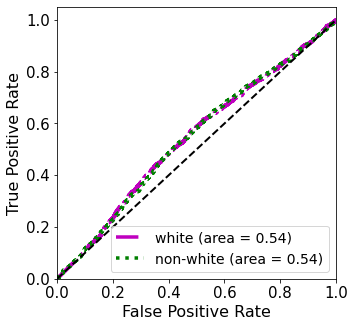} 
    \end{minipage}
    \vfill
    \begin{minipage}[t]{.45\columnwidth}
    \centering
    \includegraphics[width=\columnwidth]{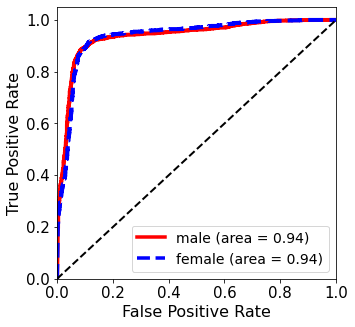}
    \end{minipage}
    \hfill
    \begin{minipage}[t]{.45\columnwidth}
    \centering
    \includegraphics[width=\columnwidth]{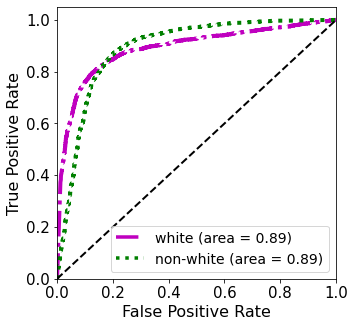} 
    \end{minipage}
    \caption{The ROC curve of (top) a \nc \model and (bottom) an \hbc \model trained by the regularized attack in \HBCP. The \pub attribute is Gender and the \priv attribute is Race (in UTKFace), and $(\beta^{y}, \beta^{s})$ in Eq~\eqref{eq:loss_func_binary} is $(.5,.5)$.}
    \label{fig:gen_race_bin}
    \end{center}
    \vspace{-.3cm} 
\end{figure}

\subsubsection{Architectures} For $\cF$, we use a DNN architecture similar to the original paper of UTKFace dataset~\cite{zhang2017age} that includes 4 convolutional layers and 2 fully-connected layers with about 250K trainable parameters. For $\cG$, we use a simple 3-layer fully-connected classifier with about 2K to 4K trainable parameters; depending on the value of $Y$. The implementation details for  $\cF$ and $\cG$ are presented in Appendix~\textcolor{blue}{\ref{apx:dnn_arch}}. For all experiments, we use a batch size of 100 images, and Adam optimizer~\cite{kingma2014adam} with learning rate $.001$. After fixing $\beta$ multipliers, we run training for 50 epochs, and choose models of the epoch that both $\cF$ and $\cG$ achieve the best trade-off for both $y$ and $s$~(based on $\beta^y$ and $\beta^s$) on the validation set, respectively. That is, the models that give us the largest $\beta^y\delta^y+\beta^s\delta^s$ on the validation set during training. Notice that, the fine-tuning is a task at training time, and a \server with enough data and computational power can find near-optimal values for $(\beta^y,\beta^s)$, as we do here using the validation set. In the following, all reported values for honesty $\delta^y$ and curiosity $\delta^s$ of \hbc \models are the accuracy of the final $\cF$ and $\cG$ on the test set. Finally, in all \NC settings, values in {\it  italic} show the effect of overlearning~\cite{song2019overlearning}, that is the accuracy of a parameterized $\cG$ in inferring a \priv attribute from a \nc \model.

\subsection{UTKFace: Gender vs. Race} 
As the first result, we present a simple result on UTKFace. We set Gender as the \pub attribute $y$ and Race~(White, Non-White) as the \priv attribute $s$. The training set includes 52\% Male and 48\% Female, where 42\% of samples are labeled as White and 58\% as Non-White. Figure~\textcolor{blue}{\ref{fig:gen_race_bin}}  shows the ROC curves for this experiment. For honesty, in the top-left plot, the \nc \model achieves $0.97$ area under the ROC curve~(AUC), whereas in the bottom-left plot the \model is \hbc but it still achieves a considerable $0.94$ AUC. For curiosity, the \nc \model in the top-right plot is not informative about Race and it basically is as good as a random guess. But, the \hbc \model in the bottom-right can achieve $0.89$ AUC on predicting Race via the regularized attack.

\subsection{CelebA: Smiling vs. Other Attributes}\label{sec:smile_vs_others}
Here, we consider a smile detection task on CelebA where $y$ is set to Smiling and $s$ is set to one of MouthOpen, Male, HeavyMakeup, or WavyHair. To better understand the results, Table~\textcolor{blue}{\ref{tab:smile_vs_others}}~(A) shows the characteristics of these attributes compared to Smiling attribute in the training dataset. As expected, MouthOpen is the most correlated one~(empirical MI is $.231$ bits) while WavyHair is the least correlated attribute~(empirical MI is $.003$ bits), to Smiling. We also show the test accuracy of $\cF$ as a \nc \model trained for each attribute. We see that Male is the easiest attribute~(97\% accuracy) and WavyHair is the most difficult one~(77\% accuracy). Table~\textcolor{blue}{\ref{tab:smile_vs_others}}~(B), (C), and (D) show how these factors, {\em correlation} and {\em easiness}, affect the performance of \hbc \models: for different attacks and trade-offs based on $\beta^y$ and $\beta^s$. Our findings are: 

\textit{1.} The outputs of a \nc binary \model, either \Rval or \Pval, do not reveal information about \priv attributes more than what one could already infer from the knowledge about underlying data distribution. Thus,  while overlearning~\cite{song2019overlearning} has shown serious problems when the \server observes a subset of internal representations, here we see that when the \server only observes the output of a binary \model, then overlearning is not a major problem (we show this in other settings as well). On the other hand, \hbc \models can effectively learn to encode the \priv attribute in a single-valued output (that is more restricted than the internal representation). 

\textit{2.} In regularized attacks in Table~\textcolor{blue}{\ref{tab:smile_vs_others}} (B), with a very small loss in $\delta^y$ (less than $1\%$), we can get $\delta^s$ very close to the accuracy that we could have achieved if we could run a separate \model for the \priv attribute.
In parameterized attacks, it is easier to encode \priv information into the \Rval output than the \Pval output. The attacks in \HBCR are highly successful in curiosity in all four cases in Table~\textcolor{blue}{\ref{tab:smile_vs_others}} (C), almost without any damage to the honesty. On the other hand, in \HBCP it is harder to establish an efficient trade-off between honesty and curiosity.  Since $\texttt{softmax}(\bx)=\texttt{softmax}(\bx+a)$ for all $a\in\bbR$, there are infinitely many vectors in $\bbR^Y$ (in \HBCR) that can be mapped into the same vector in $\bbP^Y$ (in \HBCP). However, what one can learn about the \priv attribute in \HBCP is still much more than \NC. In the following, we will see that when $Y>2$, parameterized attack in \HBCP is also very successful.

\textit{3.} The easiness of the \priv attribute plays an important role. For example, in face image processing, gender classification is in general an easier task than detecting heavy makeup (as it might be easier for human beings as well). Therefore, while MI between Smiling and HeavyMakeup is larger than Smiling and Male, the curiosity in inferring Male attribute is more successful than HeavyMakeup. Moreover, while (due to MI) $\delta^s$ of MouthOpen is about 11\% more than Male in \NC, in contrast to overlearning attack, the correlation is not that important in \hbc settings compared to the easiness, as we see that for Male all attacks are as successful as MouthOpen. For the same reason, for attribute WavyHair it is more difficult to achieve high curiosity as it is not an easy task. It is worth noting that, for difficult attributes we may be able to improve the curiosity by optimizing $\cF$ for that specific attribute, \eg through neural architecture search. But for fair comparisons, we use the same DNN for all experiments and we leave the architectures optimization for \hbc \models to future studies.

\begin{table}[t] 
\centering
\caption{Results where the \pub attribute is HairColor and the \priv attribute is (A) Male, (B) Smiling, or (C) Attractive. Here $Y$=$3$, $S$=$2$, and $\beta^x$=$0$. Values are in percentage~(\%).}
\label{tab:hair_vs_others_celeba} 
{%
\begin{tabular}{ccccc}
Setting &  Attack  & ($\beta^{y}$, $\beta^{s})$ & $\delta^y$   & $\delta^s$\\ 
\toprule
\multicolumn{5}{c}{\textbf{(A)} Male (class distribution 0:66\%, 1:34\%)}\\
\toprule
\multirow{1}{*}{\NC} & \multirow{1}{*}{\Rval\cite{song2019overlearning}} & $(1.,\textit{.0})$ & $92.94\pm.58$ & $\mathit{75.86\pm.03}$ \\ \cline{2-5}  
\multirow{1}{*}{\HBCR} & \multirow{1}{*}{Parameterized} & $ (.7,.3)$ & $92.85\pm.42$ & $97.27\pm.09$\\\cline{2-5} 
\multirow{6}{*}{\HBCP} & \multirow{3}{*}{Parameterized}  
& $ (.7,.3)$ &  $92.12\pm.27$  & $94.23\pm1.4$\\
&& $(.5,.5)$&  $91.98\pm.12$ & $96.87\pm.13$ \\
&& $(.3,.7)$& $67.63\pm.25$ & $97.52\pm.08$ 
\\\cline{3-5} 
& \multirow{3}{*}{Regularized}
&$(.7,.3)$ &  $92.79\pm.22$ & $95.11\pm.22$\\
&&$(.5,.5)$&  $92.78\pm.28$ & $96.68\pm.12$ \\
&&$(.3,.7)$&  $92.96\pm.33$ & $97.02\pm.04$ \\
\bottomrule
\multicolumn{5}{c}{\textbf{(B)} Smiling (class distribution 0:49\%, 1:51\%)} \\
\toprule
\multirow{1}{*}{\NC} & \multirow{1}{*}{\Rval\cite{song2019overlearning}} & $( 1.,\textit{.0})$ & $92.94\pm.58$ & $\mathit{56.12\pm.21}$\\ \cline{2-5} 
\multirow{1}{*}{\HBCR} & \multirow{1}{*}{Parameterized} & $ (.7,.3)$ & $92.56\pm.46$ & $91.79\pm.12$ \\\cline{2-5}
\multirow{6}{*}{\HBCP} & \multirow{3}{*}{Parameterized}  
& $ (.7,.3)$ & $92.96\pm.44$  & $88.59\pm1.8$ \\
&& $(.5,.5)$& $91.27\pm.35$ & $89.18\pm.06$ \\
&& $(.3,.7)$& $88.05\pm1.3$  & $91.07\pm.51$\\\cline{3-5}
& \multirow{3}{*}{Regularized}
&$(.7,.3)$ & $92.80\pm.28$ & $86.22\pm.38$ \\
&&$(.5,.5)$& $92.52\pm.43$ & $90.67\pm.15$ \\
&&$(.3,.7)$&  $92.31\pm.32$ & $91.59\pm.07$ \\ 
\bottomrule
\multicolumn{5}{c}{\textbf{(C)} Attractive (class distribution 0:40\%, 1:60\%)}\\
\toprule
\multirow{1}{*}{\NC} & \multirow{1}{*}{\Rval\cite{song2019overlearning}} & $(1.,\textit{.0})$ & $92.94\pm.58$ & $\mathit{60.73\pm.21}$ \\\cline{2-5}  
\multirow{1}{*}{\HBCR} & \multirow{1}{*}{Parameterized} & $ (.7,.3)$ &  $92.81\pm.16$ & $76.97\pm.07$  \\\cline{2-5} 
\multirow{6}{*}{\HBCP} & \multirow{3}{*}{Parameterized}  
& $ (.7,.3)$ &  $93.03\pm.33$ & $71.54\pm.60$ \\
&& $(.5,.5)$& $92.62\pm.64$ & $76.18\pm.12$ \\
&& $(.3,.7)$& $92.66\pm.33$  & $75.57\pm.37$ \\\cline{3-5} 
& \multirow{3}{*}{Regularized}
&$(.7,.3)$ & $91.95\pm.35$ & $69.06\pm.39$ \\
&&$(.5,.5)$&  $92.33\pm.44$ & $74.38\pm.13$ \\
&&$(.3,.7)$&  $91.09\pm.46$ & $75.70\pm.27$ \\ 
\bottomrule
\end{tabular}%
}
\end{table}

\subsection{CelebA: HairColor vs. Other Attributes}
Moving beyond binary \models, in Table~\textcolor{blue}{\ref{tab:hair_vs_others_celeba}} we present a use-case of a three-class classifier (Y=3) for the \pub attribute of HairColor, where there are $40\%$ samples of BlackHair, $34\%$ BrownHair,  and $26\%$ BlondHair. We consider three \priv attributes with different degrees of easiness: (A) Male, (B) Smiling, and (C) Attractive. While in \NC setting, attacks on overlearning~\cite{song2019overlearning} are not very successful (even when  releasing the \Rval outputs), our parameterized attacks, in \HBCR, are very successful without any meaningful damage to the honesty of \model.  In \HBCP, while it is again harder to train a parameterized attack as successful as \HBCR, we do find successful trade-offs if we fine-tune $(\beta^y,\beta^s)$; particularly for the regularized attack.

\subsection{UTKFace: Sensitive Attributes with S>2} 
\begin{table*}[t]
\caption{The honesty  ($\delta^y$) and curiosity  ($\delta^s$) of an \hbc \model trained via parameterized attack, where the \pub attribute ($y$) is Age and the \priv attribute ($s$) is Race. Values are in percentage (\%)}
\label{tab:age_race_utk}
\resizebox{\textwidth}{!}{
\begin{tabular}{ccccc|cc|cc|cc}
 & &  & \multicolumn{2}{c}{$S=2$} & \multicolumn{2}{c}{$S=3$} & \multicolumn{2}{c}{$S=4$} & \multicolumn{2}{c}{$S=5$} \\\cline{2-11}  
              &   & $(\beta^y,\beta^s)$ & $\delta^y$   & $\delta^s$   & $\delta^y$   & $\delta^s$   & $\delta^y$   & $\delta^s$   & $\delta^y$   & $\delta^s$   \\\cline{3-11}
\multirow{3}{*}{$Y=2$} & $\mathsf{NC}$ & (1.,\textit{0.}) & $83.03\pm.33$ & $\mathit{62.44\pm.97}$ & $83.03\pm.33$ & $\mathit{52.34\pm1.4}$ & $83.03\pm.33$ & $\mathit{45.74\pm.13}$ & $83.03\pm.33$ & $\mathit{44.67\pm.17}$ \\
     & \multirow{1}{*}{\HBCR} & (.7,.3) & $83.60\pm.22$  & $86.77\pm.11$   &  $83.22\pm.27$  &  $80.06\pm.88$  &   $83.10\pm.18$  &  $74.12\pm.33$   &  $83.07\pm.13$  &  $72.52\pm.34$ \\
    & \multirow{2}{*}{\HBCP} & (.7,.3) & 
                    $82.87\pm.21$  &  $69.89\pm.40$  &   $82.88\pm.27$  &  $63.28\pm.30$  &    $81.67\pm.54$  & $55.01\pm.57$      &  $81.94\pm.81$  &  $53.84\pm.93$  \\ 
                   &  & (.5,.5) & $68.86\pm.93$ & $84.75\pm1.1$ & $64.60\pm1.0$ & $82.88\pm.56$  & $72.28\pm.65$ & $62.87\pm.29$ & $65.71\pm.86$ & $71.54\pm1.6$ \\ \cline{3-11}  
\multirow{3}{*}{$Y=3$} & $\mathsf{NC}$ & (1.,\textit{0.}) & $81.09\pm.37$ & $\mathit{65.99\pm.40}$ & $81.09\pm.37$ & $\mathit{56.37\pm.47}$ & $81.09\pm.37$ & $\mathit{48.28\pm.49}$ & $81.09\pm.37$ & $\mathit{46.67\pm.41}$ \\ 
&\multirow{1}{*}{\HBCR} & (.7,.3) & $81.67\pm.33$  &  $86.35\pm.65$    & $81.35\pm.28$ &  $83.42\pm.29$  & $81.23\pm.29$  &  $76.62\pm.56$ &  $81.30\pm.28$  &  $76.24\pm.59$ \\
                   & \multirow{2}{*}{\HBCP} & (.7,.3) & 
                   $81.19\pm.02$   &  $79.07\pm.15$  &    $80.76\pm.36$  &  $72.90\pm.23$  &   $80.10\pm.32$  &  $66.40\pm1.1$      &  $80.29\pm.07$  &  $67.74\pm.30$  \\
                  &   & (.5,.5) & $78.17\pm.38$ & $86.17\pm.20$ & $69.56\pm.75$ & $80.14\pm.30$ & $69.03\pm.60$ & $76.63\pm.19$ & $68.54\pm.98$ & $76.69\pm.72$ \\  \cline{3-11} 
\multirow{3}{*}{$Y=4$} & $\mathsf{NC}$ & (1.,\textit{0.}) & $68.59\pm.24$ & $\mathit{66.93\pm.39}$ & $68.59\pm.24$ & $\mathit{58.02\pm.55}$ & $68.59\pm.24$ & $\mathit{49.23\pm.53}$ & $68.59\pm.24$ & $\mathit{41.13\pm.15}$ \\
& \multirow{1}{*}{\HBCR} & (.7,.3) & $68.59\pm.27$  &  $86.91\pm.27$   &  $68.59\pm.49$  &  $84.17\pm.26$  &  $68.29\pm.26$  &  $79.46\pm.16$   &  $68.40\pm.50$  &  $78.79\pm.13$ \\
                   & \multirow{2}{*}{\HBCP} & (.7,.3) &  
                   $68.15\pm.42$  &  $78.10\pm.32$  &     $67.45\pm.25$  &  $74.70\pm.43$     &    $66.30\pm.22$  &  $69.01\pm1.3$  & $65.70\pm.51$  &  $69.37\pm.28$   \\
                   &  & (.5,.5) & $64.48\pm.50$ & $86.68\pm.11$ & $62.69\pm.19$ & $84.11\pm.51$ & $58.95\pm.77$ & $77.48\pm.53$ & $58.19\pm.57$ & $76.89\pm.99$ \\ \cline{3-11} 
\multirow{3}{*}{$Y=5$} & $\mathsf{NC}$ & (1.,\textit{0.}) & $62.24\pm.61$ & $\mathit{67.00\pm.32}$ & $62.24\pm.61$ & $\mathit{58.69\pm.44}$ & $67.00\pm.32$ & $\mathit{50.37\pm.35}$ & $67.00\pm.32$ & $\mathit{49.22\pm.62}$ \\
& \multirow{1}{*}{\HBCR} & (.7,.3) &  $62.72\pm.20$  &  $86.63\pm.05$   &   $62.46\pm.49$  &  $83.79\pm.16$   &  $61.84\pm.77$   &   $78.53\pm.80$   &  $62.40\pm.31$  &  $78.28\pm.20$ \\
                   & \multirow{2}{*}{\HBCP} & (.7,.3)
                   & $62.08\pm.14$  &  $79.41\pm.75$   &      $61.35\pm.58$  &  $74.57\pm.91$     &     $60.90\pm.39$  &  $69.32\pm.49$     & $60.70\pm.52$ &  $69.56\pm1.2$  \\
                   &  & (.5,.5) & $58.53\pm.85$ & $86.88\pm.34$ & $56.63\pm.79$ & $84.06\pm.09$ & $53.27\pm.14$ & $77.45\pm.11$ & $54.61\pm.20$ & $77.30\pm1.0$ \\ \bottomrule
\end{tabular}%
} 
\end{table*}
\begin{table*}[t]
\centering
\caption{The honesty  ($\delta^y$) and curiosity  ($\delta^s$) of an \hbc \model trained via different  attacks, where the \pub attribute ($y$) is Race and the \priv attribute ($s$) is Gender. Values are in percentage (\%).}
\label{tab:gender_race_utk}
\resizebox{\textwidth}{!}{%
\begin{tabular}{cccccc|cc|cc|cc}
& & &  & \multicolumn{2}{c}{$Y=2$} & \multicolumn{2}{c}{$Y=3$} & \multicolumn{2}{c}{$Y=4$} & \multicolumn{2}{c}{$Y=5$} \\\cline{5-12}
&Setting &  Attack  & $(\beta^y, \beta^s)$  & $\delta^y$   & $\delta^s$   & $\delta^y$   & $\delta^s$   & $\delta^y$   & $\delta^s$   & $\delta^y$   & $\delta^s$   \\\cline{2-12}
\multirow{5}{*}{$S=2$}  & \NC & \Rval\cite{song2019overlearning} & (1.,\textit{0.}) & $87.67\pm.49$ & $\mathit{56.11\pm.44}$  & $85.50\pm.13$ & $\mathit{56.04\pm2.4}$  & $81.15\pm.23$ & $\mathit{58.91\pm.74}$ & $80.69\pm.23$ & $\mathit{61.06\pm2.0}$ \\\cline{3-12}
                  & \HBCR & Parameterized & (.7, .3) & $88.10\pm.22$ & $89.36\pm.03$ & $85.77\pm.30$ & $89.14\pm.21$ & $82.05\pm.14$ & $89.19\pm.33$ & $81.43\pm.34$ & $89.00\pm.42$ \\\cline{3-12}  
                  &\multirow{5}{*}{\HBCP}& \multirow{2}{*}{Parameterized} &
                  (.7, .3) & $87.43\pm.09$ & $51.66\pm.00$ & $85.71\pm.16$ & $83.51\pm1.1$ & $81.10\pm.23$ & $83.60\pm.22$ & $80.46\pm.26$ & $88.46\pm.18$ \\
                  &&& (.5, .5) & 
                  $83.67\pm.27$ & $80.17\pm2.1$ & $82.89\pm.59$ & $85.69\pm.54$ & $70.36\pm1.5$ & $86.05\pm1.6$ & $77.40\pm.35$ & $88.36\pm.45$ \\
                  \cline{4-12}  
                   &&\multirow{2}{*}{Regularized} & (.7, .3) &  $87.30\pm.29$  &  $75.29\pm.89$ & $84.89\pm.20$  &   $77.53\pm.33$ & $80.64\pm.31$  &  $76.29\pm.45$ & $80.24\pm.09$  &  $78.01\pm.62$\\
                   && & (.5, .5) & $86.61\pm.15$  &  $83.23\pm.27$ & $83.89\pm.44$  &  $86.61\pm.61$ & $79.03\pm.21$ &  $86.29\pm.32$ & $78.61\pm.65$  &  $85.97\pm.28$\\
                   \hline
\end{tabular}%
}
\end{table*}   
To evaluate tasks with $S>2$, we provide results of several experiments performed on UTKFace in Table~\textcolor{blue}{\ref{tab:age_race_utk}} and Table~\textcolor{blue}{\ref{tab:gender_race_utk}} (also some complementary results in Appendix~\textcolor{blue}{\ref{appx:add_res}}, Table~\textcolor{blue}{\ref{tab:model_cap_utk_33}}, Tables~\textcolor{blue}{\ref{tab:race_age_utk}}, and Table~\textcolor{blue}{\ref{tab:race_gender_utk}}). In each experiment, we set one of Gender, Age, or Race, as $y$ and another one as $s$, and compare the achieved $\delta^y$ and $\delta^s$. See Appendix~\textcolor{blue}{\ref{apx:utk}} for the details of how we created labels for different values of $Y$ and $S$. 
Our findings are:
 
\textit{1.} An \hbc \model can be as honest as a \nc \model while also achieving a considerable curiosity. For all \HBCR cases, the $\delta^y$ of an \hbc \model is very close to $\delta^y$ of a corresponding \model in \NC. Moreover, we see that in some situations, making a \model \hbc even helps in achieving a better generalization and consequently getting a slightly better honesty; which is very important as an \hbc \model can look as honest as possible (we elaborate more on the cause of this observation in Appendix~\textcolor{blue}{\ref{sec:generalization}}).

\textit{2.} When having access to \Rval outputs, the attack is highly successful in all tasks, and in many cases, we can achieve similar accuracy to a situation where we could train $\cF$ for that specific \priv attribute. For example, in Table~\textcolor{blue}{\ref{tab:age_race_utk}} for $S=3$ and $Y>2$, we can achieve about $83\%$ curiosity in inferring the Race attribute from a \model trained for Age attribute. When looking at Table~\textcolor{blue}{\ref{tab:race_age_utk}} where Race is the \pub attribute, we see that the best accuracy a \nc \model can achieve for Race classification is about $85\%$.

\begin{table*}[t]
\caption{The effect of entropy minimization on the UTKFace dataset for the \pub attribute Age and the \priv attribute Race where $Y=S=3$. We also show the average of the entropy of \model's output by  $\widetilde{\tH}(\byh)$ in bits.} 
\label{tab:entropy_betax_utk_33}
\resizebox{\textwidth}{!}{%
\begin{tabular}{ccccc|
ccc|ccc|ccc}
 & 
  $(\beta^y, \beta^s)$ &
  \multicolumn{3}{c|}{$\beta^x = .0$} & 
  \multicolumn{3}{c|}{$\beta^x = .2$} &
  \multicolumn{3}{c|}{$\beta^x = .4$} &
  \multicolumn{3}{c}{$\beta^x = .8$} \\ 
\multicolumn{1}{c}{} &
  \multicolumn{1}{c}{} &
  $\delta^y$ & $\delta^s$ & $\widetilde{\tH}(\byh)$ & 
  $\delta^y$ &  $\delta^s$ & $\widetilde{\tH}(\byh)$ & $\delta^y$ & $\delta^s$ & $\widetilde{\tH}(\byh)$ & $\delta^y$ & $\delta^s$ &  $\widetilde{\tH}(\byh)$ \\\cline{3-14}  
\NC                    & (1., \textit{.0}) 
& 81.43$\pm$.37  & \textit{58.14$\pm$.06} & .48$\pm$.05 & 
81.04$\pm$.09   & \textit{56.13$\pm$.14} & .45$\pm$.06 & 80.90$\pm$.27  & \textit{58.60$\pm$.07} & .40$\pm$.05 & 81.15$\pm$.41    & \textit{55.94$\pm$.26} & .32$\pm$.02 \\\cline{3-14} 
\multirow{2}{*}{\HBCR} & (.7, .3)  
& 81.32$\pm$.31  &  82.97$\pm$.27  &  .63$\pm$.05 & 
81.59$\pm$.42  &  82.90$\pm$.24  &  .45$\pm$.02 &  81.62$\pm$.37  &  81.98$\pm$.77  &  .36$\pm$.02 &  81.61$\pm$.31  &  81.40$\pm$.41  &  .28$\pm$.03 \\
& (.5, .5)  
&  80.23$\pm$.42  &  84.82$\pm$.35  &  .76$\pm$.03 &
80.60$\pm$.14  &  84.60$\pm$.32  &  .54$\pm$.03 & 80.95$\pm$.41  &  83.86$\pm$.54  &  .39$\pm$.01 &  81.26$\pm$.22  &  83.92$\pm$.32  &  .26$\pm$.01 \\ \cline{3-14}
\multirow{2}{*}{\HBCP} & (.7, .3)  
& 81.04$\pm$.26  &  73.34$\pm$.29  &  .72$\pm$.02 & 
81.26$\pm$.28  &  69.77$\pm$.50  &  .51$\pm$.02 &  81.01$\pm$.55  &  64.54$\pm$2.5  &  .37$\pm$.04 &  81.01$\pm$.46  &  56.28$\pm$.51  &  .27$\pm$.05 \\ 
& (.5, .5) 
& 69.12$\pm$.91  &  80.81$\pm$.97  &  1.05$\pm$.01  &  
76.20$\pm$.29  &  76.27$\pm$.37  &  .76$\pm$.01 & 80.13$\pm$.15  &  74.01$\pm$.46  &  .58$\pm$.01 &  80.59$\pm$.16  &  70.28$\pm$.15  &  .33$\pm$.02 \\\bottomrule  
\end{tabular}%
} 
\end{table*}

\textit{3.} In \HBCP, it is more challenging to achieve a high curiosity via a parameterized attack, unless we sacrifice more honesty. Particularly for tasks with $S>Y$, where the \priv attribute is more granular than the \pub attribute.  Also, while we observed successful regularized attacks for \HBCP in Table~\textcolor{blue}{\ref{tab:smile_vs_others}}~(C), regularized attacks cannot be applied to tasks with $S>2$. Yet, even in this case of having only access to \Pval outputs, the curiosity is much higher than what can be learned from the \Rval output of a \nc \model (via overlearning attack). Although the curiosity in \HBCR is more successful than \HBCP,  the difference between these two gets smaller as the size of output $Y$ gets larger. 

\textit{4.} Attacks are highly successful when $S \leq Y$, as there is more capacity in the released output. However, the attack is successful in scenarios when $S > Y$ as well. The most challenging case is where $Y=2$ and when we only have access to the \Pval outputs, because in these tasks we only release one value (\ie $\yh_1 = 1-\yh_2$). Moreover, we see in Table~\textcolor{blue}{\ref{tab:gender_race_utk}} that for \HBCP with $S=Y=2$, regularized attacks achieve much better trade-offs than parameterized attacks.

\subsection{Entropy Minimization with $\beta^x>0$}

We examine the entropy minimization~(\ie compression) of the \model's outputs and its effect on the achieved trade-off for honesty and curiosity.  Table~\textcolor{blue}{\ref{tab:entropy_betax_utk_33}}  compares the results of different values chosen for $\beta^x$ in Eq~\eqref{eq:optim_F_2_emp} (see Figure~\textcolor{blue}{\ref{fig:adv_train}}). An important observation is that compression is mostly helpful to the honesty. This is expected, due to our discussion in Section~\textcolor{blue}{\ref{sec:general_cat}}, and findings in previous related works~\cite{tishby2015deep, vera2018role}. Moreover, we see that compression is more effective in improving the honesty of the \model in situations where we assign more weight to the curiosity of the \model.

\begin{figure}[t] 
    \centering
    \includegraphics[width=.78\columnwidth]{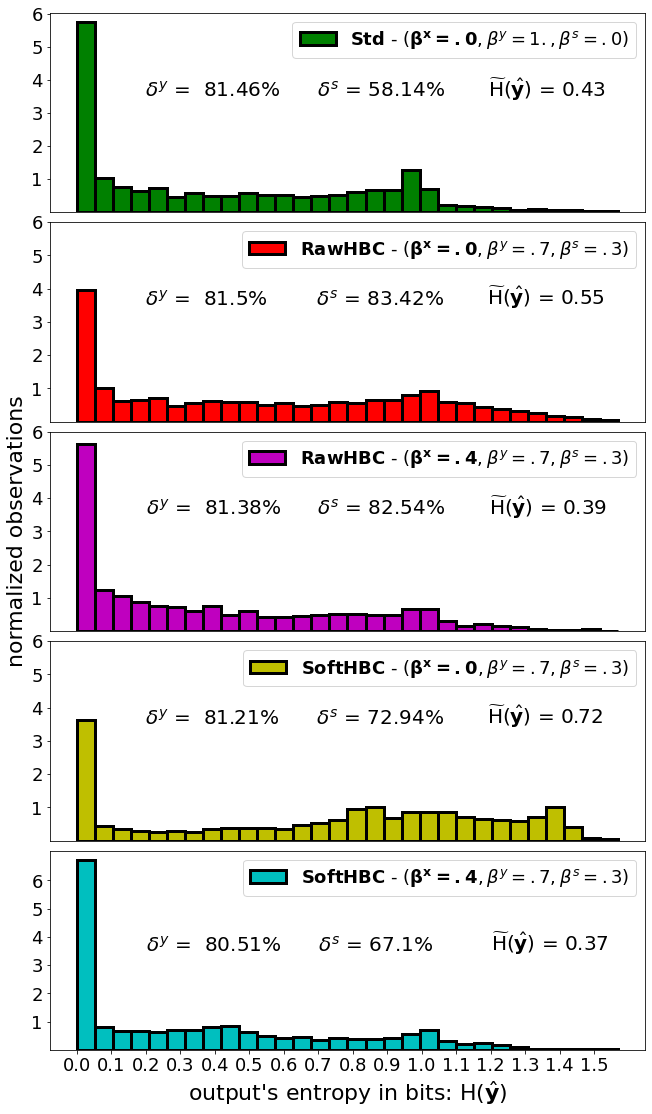} 
    \vspace{-.3cm} 
    \caption{The normalized density histogram (\ie number of observations) of the output entropies of \model with and without compression.  Here, $y$ is Age and $s$ is Race with $Y=S=3$ on UTKFace. From top to bottom, \model is trained (1) \nc, (2, 3) \HBCR where we release the \Rval outputs, and  (4, 5) \HBCP where we release the \Pval outputs. For (2) and (4) we set $\beta^{x}=.0$, and for (3) and (5) we set $\beta^{x}=.4$.}  
    \label{fig:ent_dist} 
    \vspace{-.3cm}
\end{figure} 

Although Table~\textcolor{blue}{\ref{tab:entropy_betax_utk_33}} shows that large compression hurts curiosity more, this is another trade-off that a \server can utilize to make the \hbc \model less suspicious. It is important to observe that the average entropy of the \model's output, shown by $\widetilde{\tH}(\byh)$, is directly related to the curiosity weight $\beta^s$. The more curious a \model is, the larger will be the entropy of the output. 
In Figure~\textcolor{blue}{\ref{fig:ent_dist}}, we plot the histogram of normalized observations of the output's entropy for the setting of $Y$=$S$=$3$ for five scenarios in Table~\textcolor{blue}{\ref{tab:entropy_betax_utk_33}}.
We see that the output of an \hbc \model tends to have larger entropy than a \nc \model. Moreover, large entropies, \ie more than $1$ bit, are more common in \HBCP than \HBCR. A reason for this is that the capacity of \Pval outputs is smaller than \Rval outputs; hence, the \model tends to take more advantage of the existing capacity. Interestingly, when we use the entropy minimization with $\beta^x > 0$, then we observe that the entropy distribution for an \hbc \model has a smaller tail (compared to  $\beta^x = 0$) and thus can become even less suspicious than the \nc \model when we release the \Rval output. Finally, in releasing the \Pval outputs, it is a bit more challenging to keep the average entropy low.

\begin{figure*}
    \begin{center}
    \begin{minipage}[t]{\columnwidth}
    \centering
    \includegraphics[width=.85\columnwidth]{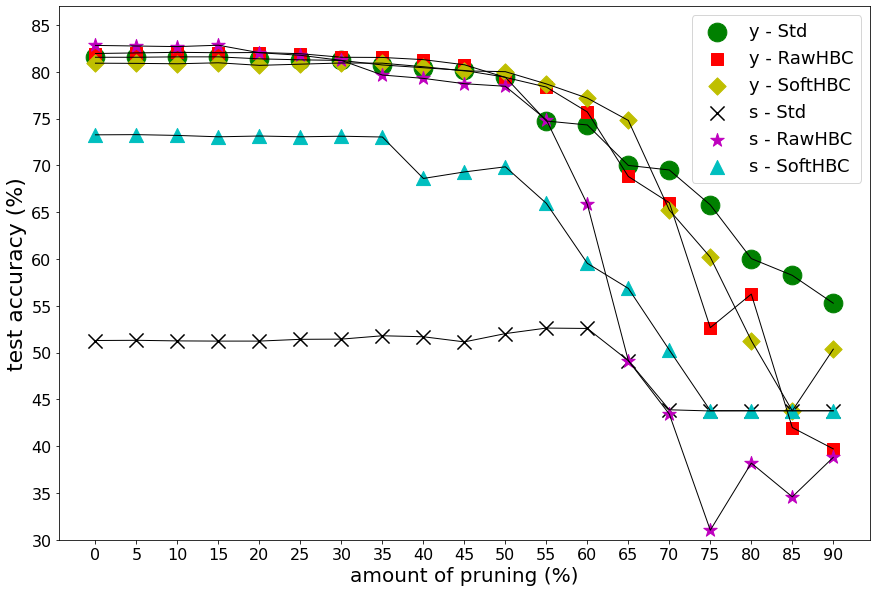}
    \end{minipage}
    \begin{minipage}[t]{\columnwidth}
    \centering
    \includegraphics[width=.85\columnwidth]{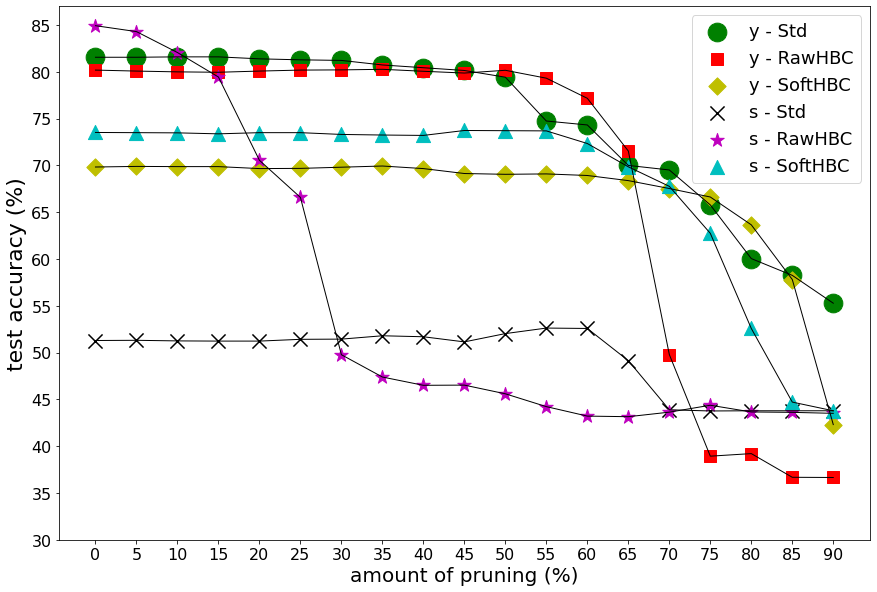}
    \end{minipage}
    \begin{minipage}[t]{\columnwidth} 
     \centering
    \includegraphics[width=.85\columnwidth]{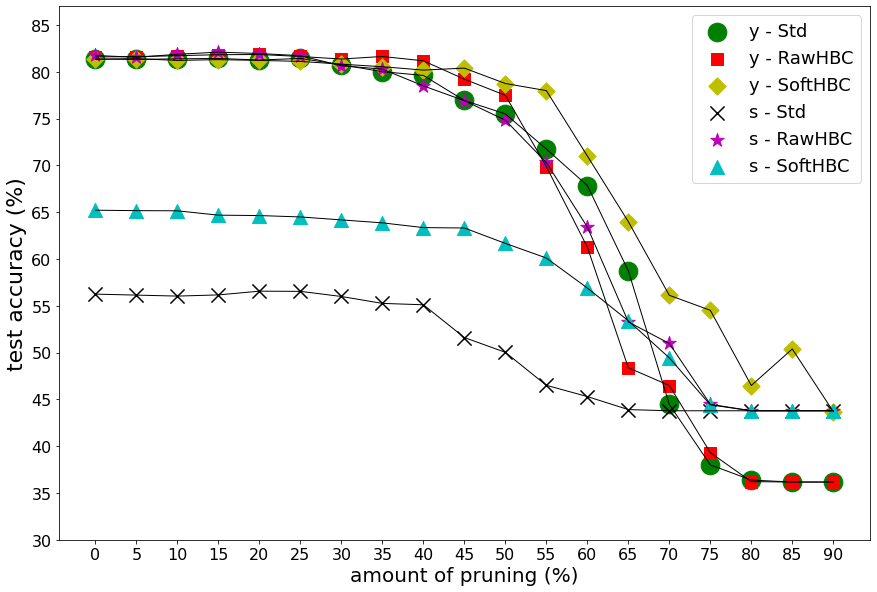}
    \end{minipage}
    \begin{minipage}[t]{\columnwidth}
    \centering
    \includegraphics[width=.85\columnwidth]{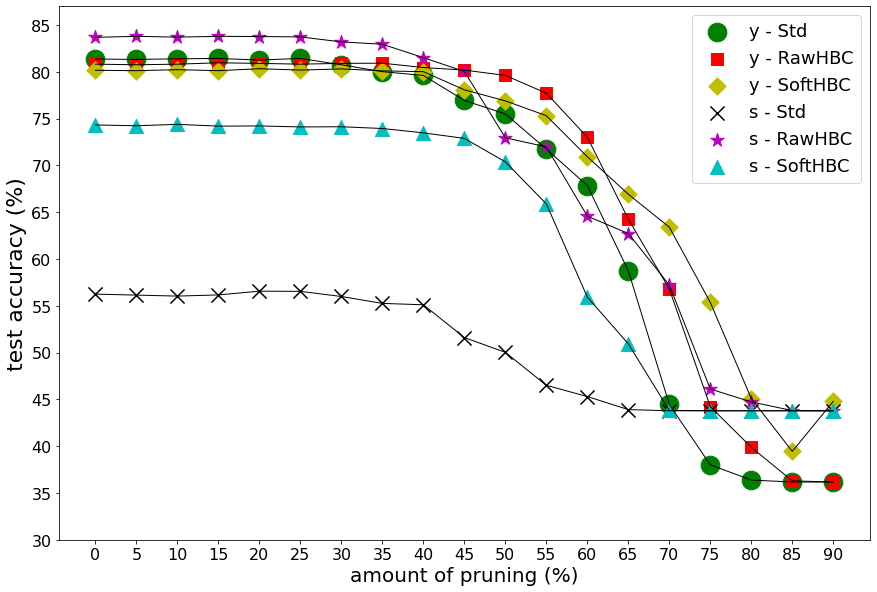}
    \end{minipage} 
    \vspace{-.3cm}
    \caption{The effect of pruning, with L1 norm, where
    the \model $\cF$ when $y$ is Age, $s$ Race, and $Y=S=3$. (Top-Left) $\beta^{x}, \beta^{y}, \beta^s=(0,.7,.3)$. (Top-Right) we set $\beta^{x}, \beta^{y}, \beta^s=(0,.5,.5)$.(Bottom-Left) $\beta^{x}, \beta^{y}, \beta^s=(.4,.7,.3)$. (Bottom-Right) we set $\beta^{x}, \beta^{y}, \beta^s=(.4,.5,.5)$.} 
    \label{fig:prune}
\end{center}
\end{figure*}

\subsection{Pruning \hbc \Models}  
As DNN \models are mostly overparameterized, a hypothesis might be that \hbc \models utilize the extra capacity of DNNs for extracting patterns that correspond to the \priv attribute. Thus, one could say that reducing a DNN's capacity, using a pruning technique~\cite{han2015deep}, might make it more difficult for a \model to be curious. For example, a \user who does not trust a \server can take the \model $\cF$ and perform some pruning technique, before making inferences, hoping that pruning will not damage the honesty but will reduce curiosity. Figure~\textcolor{blue}{\ref{fig:prune}} shows the honesty-curiosity trade-off for different amounts of pruning. As the pruning technique, we use {\em L1-Unstructured} implemented in \cite{pytorch2021prune}, where parameters with the lowest L1-norm are set to zero at inference time (\ie no re-training).   

\textit{1.} Comparing the top plots (where $\beta^x=0$) with bottom plots (where $\beta^x=0.4$) in Figure~\textcolor{blue}{\ref{fig:prune}}, we see that \models without compression show more tolerance to a large amount of pruning (>60\%) than \models with compression. This might be due to the additional constraint that we put on \models with compression.

\textit{2.} By pruning less than 50\% of the parameters, there is no significant drop in the honesty in both \nc and \hbc \models. However, for the curiosity, \hbc \models show different behaviors in different settings. When there is no compression~($\beta^x=0$), the curiosity  of \HBCR shows faster and much larger drops, compared to \HBCP. When there is compression~($\beta^x=0.4$), then drops are not different across these settings. 

\textit{3.} If we prune more than 50\% of the parameters, the drop in accuracy for both the \pub and \priv attributes are significant. However, \HBCP settings interestingly show better performances than \HBCR, for both \pub and \priv attributes. While \HBCR has usually shown better performance in previous sections, in this specific case \HBCP performs interestingly very well. 

In sum, although Figure~\textcolor{blue}{\ref{fig:prune}} shows that pruning can damage curiosity more than honesty in settings without compression. We cannot guarantee that pruning at inference time is an effective defense against \hbc models, as adding compression constraint will help the \server to make \hbc models more tolerable to pruning, and with the amount of pruning up to 50\% the curiosity remains high.

\subsection{Transferring Curiosity via \hbc Teachers}
Transferring knowledge from a {\em teacher} \model, trained on a large labeled data, to a {\em student} \model, that has access only to a (small) unlabeled data, is known as ``knowledge distillation''~(KD)~\cite{hinton2015distilling, gupta2016cross}. Since KD allows to keep sensitive data private, it has found some applications in privacy-preserving ML~\cite{papernot2016semi, wang2019private}. Let us consider a \user that owns a private unlabeled dataset and wants to train a \model on this dataset, and a \server that provides a teacher \model trained on a large labeled dataset. A common technique in KD is to force the student to mimic the teacher's behavior by minimizing the KL-divergence between the teacher's \Pval outputs, $\byh^{Teacher}$, and student's \Pval outputs,  $\byh^{Student}$~\cite{hinton2015distilling}: 
\begin{equation}
    \cL^{KL}= \sum_{i=1}^{Y} \byh^{Teacher}_{i}\log\Big({\byh^{Teacher}_{i}}/{\byh^{Student}_{i}}\Big).
\end{equation}
We show that if the teacher is \hbc, then the student trained using $\cL^{KL}$ can also become \hbc. We run experiments, similar to Section~\textcolor{blue}{\ref{sec:smile_vs_others}}, by assigning 80\% of the training set (with both labels) to the \server, and 20\% of the training set (without any labels) to the \user. At the \server side, we train an \hbc teacher via both regularized and parameterized attacks, and then at the \user side, we use the trained teacher \model to train a student only via $\cL^{KL}$ and the \user's unlabeled dataset. We also set the student's DNN to be half the size of the teacher's in terms of the number of trainable parameters in each layer. We set $(\beta^y,\beta^s)= (.5,.5)$ and consider the \HBCP setting, because $\cL^{KL}$ is based on mimicking the teacher's \Pval outputs. 

\begin{table}[]
\caption{An \hbc teacher can train an \hbc student via KD. The \pub attribute is Smiling in CelebA and the setting is \HBCP with $(\beta^x, \beta^y,\beta^s)= (.0,.5,.5)$.}
\label{tab:teach_stu}
\resizebox{\columnwidth}{!}{%
\begin{tabular}{@{}lccc|cc@{}}
 &  & \multicolumn{2}{c}{Parameterized} & \multicolumn{2}{c}{Rgularized} \\
 \multicolumn{1}{c}{$s$} & \model & $\delta^y$ & \multicolumn{1}{c}{$\delta^s$} & $\delta^y$ & $\delta^s$ \\\hline
\multirow{1}{*}{Mouth} & Teacher & $90.04\pm.61$ & $89.14\pm.68$ & $91.61\pm.24$ & $89.01\pm.34$   \\
 Open & Student & $90.30\pm.47$ & $88.07\pm.65$ & $91.29\pm.11$  & $83.69\pm.38$ \\ \cmidrule(l){2-6} 
\multirow{2}{*}{Male} & Teacher & $90.17\pm.66$ & $92.73\pm.76$ & $91.57\pm.16$ & $94.11\pm.21$ \\
 & Student & $90.20\pm.51$ & $91.62\pm.73$ & $91.01\pm.19$ & $85.71\pm.46$ \\ \cmidrule(l){2-6} 
\multirow{1}{*}{Heavy} & Teacher & $90.73\pm1.1$ & $85.17\pm.97$ & $91.56\pm.21$ & $85.58\pm.27$ \\
 Makeup & Student & $91.00\pm.43$ & $82.62\pm.49$
 & $91.23\pm.15$ &  $81.05\pm.15$ \\ \cmidrule(l){2-6}  
\multirow{1}{*}{Wavy} & Teacher & $91.99\pm.21$ & $61.05\pm1.8$ &  $91.86\pm.08$ & $71.12\pm1.2$ \\
Hair & Student & $91.78\pm.17$ & $59.72\pm1.1$ & $91.60\pm.15$ & $68.61\pm.96$ \\ \cmidrule(l){1-6} 
\end{tabular}%
}
\end{table} 

Table~\textcolor{blue}{\ref{tab:teach_stu}} shows that KD works well for the honesty in both cases; showing that an \hbc teacher can look very honest in transferring knowledge of the \pub attribute. For the curiosity, the parameterized attacks transfer the knowledge very well and are usually better than regularized attacks. A reason might be that for regularized attacks the chosen threshold $\tau$ for the teacher \model is not the best choice for replicating the attack in its students. Though, for WavyHair as an uncorrelated and difficult attribute, the regularized attack achieves a better result. 

Overall, this capability of transferring curiosity shows another risk of \hbc \models provided by semi-trusted \servers, especially that such a teacher-student approach has shown successful applications in semi-supervised learning~\cite{reed2014training, tarvainen2017mean, iscen2019label, xie2020self}. We note that there are several aspects, such as the teacher-student data ratio and distribution shift, or different choices of the loss function, that may either improve or mitigate this attack. For example, the entropy of teacher's output in KD can be controlled via a ``temperature'' parameter $t$ inside the \texttt{softmax}, \ie $\byh^{\sfP}_i = e^{\byh^{\sfR}_i /t}/\sum_{j=0}^{Y-1} e^{\byh^{\sfR}_j /t}$, which might further affect the achieved honesty-curiosity. We leave further investigation into these aspects of KD for future studies.

\section{Discussions and Related Work}
We discuss related work and potential proactive defenses against the vulnerabilities highlighted in this paper. 
\subsection{Proactive Investigation}
The right to information privacy is long known as ``the right to select what personal information about me is known to what people''~\cite{westin1968privacy}. The surge of applying ML to (almost) all tasks in our everyday lives has brought attention to the ethical aspects of ML~\cite{kearns2019ethical}. To reconstruct the \users' face images from their recordings of speech~\cite{oh2019speech2face, wen2019face} raised the concern of ``tying the identity to biology'' and categorizing people into gender or sexual orientation groups that they do not fit well~\cite{hutson2021who}. The estimation of ethnicity~\cite{wang2019expression} or detecting sexual orientation~\cite{wang2018deep} from facial images has risen concerns about misusing such ML models by adversaries that seek to determine people of minority groups. It is shown in~\cite{obermeyer2019dissecting} that a widely used healthcare model exhibits significant racial bias, causing Black patients to receive less medical care than others. The costs and potential risks associated with large-scale language models, such as discriminatory biases, are discussed in~\cite{bender2021dangers} and the research community is encouraged to consider the impacts of the ever-increasing size of DNNs beyond just the model's accuracy for a \pub task. 

In this paper, we showed another major concern regarding DNNs that enables attacks on the \users' privacy; even when users might think it is safe to only release a very narrow result of their private data. An important concern on the potential misuse of ML models is that unlike the discovery of software vulnerabilities that can be quickly patched, it is very difficult to propose effective defenses against harmful consequences of ML models~\cite{shevlane2020offense}.  It is suggested~\cite{kearns2020ethical} that tech regulators should become ``proactive'', rather than being ``reactive'', and design controlled, confidential, and automated experiments with black-box access to ML services.  While service providers may argue that ML models are proprietary resources, it is not unreasonable to allow appropriate regulators to have controlled and black-box access for regulatory purposes. For instance, a regulator can check the curiosity of ML \models provided by cloud APIs via a test set that includes samples with \priv attributes.  

\subsection{Attacks in Machine Learning}  
ML models can leak detailed \priv information about their training datasets in both white-box and black-box views~\cite{song2017machine}. Since DNNs tend to learn as many features as they can, and some of these features are inherently useful in inferring more than one attribute, then DNNs trained for seemingly non-sensitive attributes can implicitly learn other potentially \priv attributes~\cite{fredrikson2015model, song2019overlearning}. While information-theoretical approaches~\cite{moyer2018invariant, wang2018not, osia2020hybrid} are proposed for training DNNs such that they do not leak \priv attributes, it is shown that the empirical implementation of these theoretical approaches cannot effectively eliminate this risk~\cite{song2019overlearning}. A technique based on transfer learning is proposed in~\cite{song2019overlearning} to ``re-purpose'' a \model trained for \pub attribute into a model for classifying a different attribute. However, the re-purposing of a \model is different from building an \hbc \model as the former does not aim to look honest and the privacy violation is due to the further use of a \model for other purposes without the consent of the training data owner.

In model extraction (\aka model stealing) attack~\cite{tramer2016stealing}, an adversary aims to build a copy of a black-box ML model, without having any prior knowledge about the model’s parameters or training data and just by having access to the \Pval predictions provided by the model.  This attack is evaluated by two objectives~\cite{jagielski2020high}: {\em test accuracy}, which measures the correctness of predictions made by the stolen model, and {\em fidelity}, which measures the similarity in predictions (even if it is wrong) between the stolen and the original model. Model extraction has shown the richness of a \model's output in reconstructing the \model itself, but in our work we show how this richness can be used for encoding \priv attributes at inference time. 

ML enables unprecedented applications, \eg automated medical diagnosis~\cite{gulshan2016development, esteva2017dermatologist}. As personal data,\eg medical images, are highly sensitive, privacy-preserving learning, such as federated learning~\cite{mcmahan2017communication}, is proposed to train DNN \models on distributed private data~\cite{truex2019hybrid,malekzadeh2021dopamine}. \Users who own sensitive data usually participate in training a DNN for a specified \pub task. Although differential privacy~\cite{dwork2014algorithmic} can protect a model from memorizing its training data~\cite{abadi2016deep, nasr2018machine}, the threat model introduced in this paper is different from the commonly studied setting in property~\cite{melis2019exploiting}, or membership~\cite{shokri2017membership,salem2018ml} inference attacks where \model is trained on a dataset including multiple \users and the \server is curious about inferring a sensitive property about \users, or the presence or absence of a target \user in the input dataset. We consider a threat to the privacy of a single user at inference time when the \server observes only outputs of a pre-trained model. 

\subsection{Defense Challenges}
Our experiments and analyses on the performance and behavior of \hbc \models imply challenges in defending against such a privacy threat. First, we observed that distinguishing \hbc models from \nc ones is not trivial, and typical \users mostly do not have the technical and computational power and resources to examine the ML services before using them. One can suggest adding random noise to the model's outputs before sharing them~\cite{luo2020feature}, but, because of corresponding utility losses, \users cannot just simply apply such randomization to every ML model they use.  Thus, we need robust mechanisms to discover \hbc models, but also entities and systems for performing such investigations.
Second, for proactive investigations, we need datasets labeled with multiple attributes, which are not always possible. Good and sufficient data is usually in the possession of ML service providers who are actually the untrusted parties in our setting. Third, \users' data might include several types of \priv attributes, and even if we aim to distinguish \hbc models we may not know which attribute an \hbc model is trained for. There might be unrecognized \priv attributes included in some of our personal data; for example, it has recently been shown~\cite{banerjee2021reading} that DNNs can be trained to predict race from chest X-rays and CT scans of  patients, in a setting where clinical experts cannot.

We believe our work serves in improving the \users' awareness about such privacy threats, and invites the community to work on efficient mechanisms as well as systems for protecting \users' privacy against such an attack.

\section{Conclusion}
We introduced and systematically studied a major vulnerability in high-capacity ML \models that are trained by semi-trusted ML service providers. We showed that deep \models can secretly encode a \priv attribute of their private input data into their public \pub outputs. Our results show that, even when \model outputs are very restricted in their form, they are still rich enough to carry information about more than one attribute. We translated this problem into an information-theoretical framework and proposed empirical methods that can efficiently implement such an attack to the privacy of \users.  We analyzed several properties of \model outputs and specifically showed that the entropy of the outputs can represent the curiosity of the model up to a certain extent.  Furthermore,  we showed that this capability can even be transferred to other \models that are trained using  such an \hbc \model. Finally, while we showed that even \Pval outputs of a  multi-class \model can be exploited for encoding \priv information, our results suggest that it is a bit safer for a \user to release \Pval outputs than \Rval outputs; without damaging the utility.

{\bf Future Work.} We suggest the following open directions for further exploration. First, rigorous techniques that can help in distinguishing \nc and \hbc \models are needed, which is in the same direction of research in understanding and interpreting DNN behaviors~\cite{lundberg2017unified}. Second, a limitation of our proposed attacks is that the sensitive attribute has to be known at training time. We suggest extending the proposed methods, or designing new methods, for encoding more than one \priv attribute in the \model's output, which, in general, will be more challenging, but not impossible. Third, to investigate the implication and effects of employing an \hbc model in collaborative/federated learning and multi-party ML applications, where some parties might not be fully trusted.  Fourth, it is of interest to understand whether a \hbc \model reveals more information about its training dataset or less compared to \nc \models. As we can imagine a setting where a \user might also be an adversary to the \server, this is a challenge for \servers that utilize private data for training \hbc \models, and looking into such scenarios will be of interest.

\begin{acks}
This work was funded by the European Research Council (ERC) through Starting Grant BEACON (no. 677854) and by the UK EPSRC (grant no. EP/T023600/1) within the CHIST-ERA program. Anastasia thanks JPMorgan Chase \& Co for the funding received through the J.P. Morgan A.I. Research Award 2019.  Views or opinions expressed herein are solely those of the authors listed. Authors thank Milad Nasr for his help in shaping the final version of the paper.

\end{acks}
\bibliographystyle{ACM-Reference-Format}
\bibliography{refs} 
\clearpage
\appendix
\section*{Appendix}
\section{The Convex Use Case}\label{sec:apx:convex}
In this section, we first present an example of an \hbc \model with a convex loss function, then we analyze the behavior of such a \model. 
 
\subsection{An Example of Convex Loss}
Let $\bx = [x_0, x_1, \ldots, x_{M-1}, 1]$ denote the data sample with \pub attribute $y\in\{0,1\}$, and \priv attribute  $s\in\{0,1\}$, and for each $i\in[M]$ we have $x_i \in \bbR$. Let us consider a binary logistic regression \model $\cF$ that is parameterized by $\Theta \in \bbR^{M+1}$:
$$
\yh = \cF(\bx) = \sigma(\Theta^{\top}\bx) = \frac{1}{1+\exp(-\Theta^{\top}\bx)}. 
$$
To make $\cF $ a honest-but-curious classifier, the \server can choose multipliers $\beta^y \in [0,1]$ and $\beta^s \in [0,1]$ such that $\beta^y+\beta^s=1$ and build the loss function
\begin{equation}\label{eq_loss_bin_log_reg_logloss}
\begin{split}
   \cL^{\cF} =  & \quad \beta^y\cL^y + \beta^s\cL^s  \\
   = & -\beta^y \big(y\log{\yh}+(1-y)\log{(1-\yh)}\big)    \\
   & - \beta^s \big(s\log{\yh}+(1-s)\log({1-\yh})\big) \\
   = & -(\beta^y y+\beta^s s)\log{\yh} \\
   & - \big(\beta^y(1-y) + \beta^s(1-s)\big)\log(1-{\yh}) \\
   = & -(\beta^y y+\beta^s s)\log{\yh} \\
   & - \big((\beta^y+\beta^s)-(\beta^y y + \beta^s s)\big)\log(1-{\yh}) \\
   = & -(\beta^y y+\beta^s s)\log{\yh} \\
   & - \big(1-(\beta^y y + \beta^s s)\big)\log(1-{\yh}) \\ 
   = & -z\log{\yh} - \big(1-z\big)\log(1-{\yh}) \\ 
\end{split} 
\end{equation}
where $z=\beta^y y+\beta^s s$. We know that: 
\begin{enumerate}
    \item If function $\cA$ is convex and $\alpha \geq 0$, then $\alpha\cA$ is convex.
    \item If two functions $\cA_1$ and $\cA_2$ are convex, then $\cA_1 + \cA_2$ is convex.
    \item Functions  $-\log{\yh}=-\log\big(\sigma(\Theta^{\top}\bx)\big)$ and $-\log{(1-\yh)}=-\log\big(1-\sigma(\Theta^{\top}\bx)\big)$, are convex \wrt the parameters of a logistic regression model~\cite{murphy2021probabilistic}.
\end{enumerate}
Thus, considering these three facts,  $\cL^{\cF}$ in Eq~\eqref{eq_loss_bin_log_reg_logloss} is convex. 

\subsection{Analysis of Convex Loss} 
Now, we analyze the situations in which loss functions for training \hbc \models $\cF$, parameterized by $\Theta$, are convex. Let the loss function for these situations be
\begin{align*}
    \cL^{\cF}&=\beta^y\cL^{y}\big(\cF(\bx; {\Theta}),y\big)+\beta^s\cL^{s}\big(\cG(\cF(\bx; {\Theta})),s\big)\\
    &=\beta^y\cL^{y}({\Theta})+\beta^s\cL^{s}(\Theta),
\end{align*}
where we assume $\cG$ is a fixed  attack, and in the second line we just simplified the equation to focus on the trainable parameters $\Theta$. Let ${\Theta}^y_*$ and ${\Theta}^s_*$ denote the optimal parameters that minimize $\cL^{\mathbf{y}}$ and $\cL^{\mathbf{s}}$, respectively. Assume that, at each iteration $t$ of training, loss functions $\cL^{\mathbf{y}}$ and $\cL^{\mathbf{s}}$ are $\mu^\mathbf{y}$- and $\mu^\mathbf{s}$-strongly convex functions with respect to ${\Theta}_t$, respectively. Considering the gradient descent dynamics in continuous time~\cite{wilson2018lyapunov},
\begin{align*}
    d{\Theta}_t = -\nabla_{\Theta}\cL^{\cF} dt,
\end{align*}
we can define the Lyapunov function $V_t=\frac{1}{2}||{\Theta}_t-{\Theta}^y_*||_2^2$. Observe that by strong convexity of $\cL^{\mathbf{y}}$, the optimum of this Lyapunov function is ${\Theta}^y_*$ and it is unique. Now, we can analyze the dynamics of this Lyapunov function to better understand the convergence of the gradient descent dynamics when the loss function is set to be optimizing both for $y$ and $s$. 

First, let us consider $\beta^s=0$, then
\begin{equation}\label{eq:convex_lyap}
    \begin{split}
     dV_t &= ({\Theta}_t-{\Theta}^y_*)^\top  d{\Theta }_t = - ({\Theta}_t-{\Theta}^y_*)^\top \nabla_{\Theta}\cL^{\cF} dt\\
    &=-\beta^y({\Theta}_t-{\Theta}^y_*)^\top\nabla_{{\Theta}}\cL^{y}({\Theta}_t)dt\\
    &=-\beta^y({\Theta}_t-{\Theta}^y_*)^\top\big(\nabla_{{\Theta}}\cL^\mathbf{y}({\Theta}_t)-\nabla_{{\Theta}}\cL^\mathbf{y}({\Theta}^y_*)\big)dt\\
    &\leq -\beta^y\mu^\mathbf{y}||{\Theta}_t-{\Theta}^y_*||_2^2dt,
    \end{split}
\end{equation}
where in the second line we use the gradient descent dynamics, in the third line we use the fact that $\nabla_{\Theta}\cL^{y}({\Theta}^y_*)=0$ (remember that we assume ${\Theta}^y_*$ is the set of parameters that is optimal for $\cL^{\mathbf{y}}$), and in the last line we use the strong convexity of the function. By integrating both sides of inequality in~(\ref{eq:convex_lyap}) we obtain 
\begin{align}
    V_t\leq e^{-t\beta^y\mu^\mathbf{y}}V_0,
\end{align}
so that as $t$ increases, we converge closer to the optimum ${\Theta}^y_*$; that is a well-known result~\cite{wilson2018lyapunov}. 

Now, we show how the addition of the loss term on the \priv attribute, $\cL^s$, prevents the convex \model from fully converging. Let $\beta^s>0$, then
\begin{align*}
    dV_t &= -({\Theta}_t-{\Theta}^y_*)^\top\nabla_{\Theta}\beta^y\cL^{y}({\Theta}_t)dt-\beta^s({\Theta}_t-{\Theta}^y_*)^\top\nabla_{{\Theta}}\cL^{s}({\Theta}_t)dt\\ 
    &\leq -\beta^y\mu^\mathbf{y}||{\Theta}_t-{\Theta}^y_*||_2^2dt + \beta^s \big(\cL^{s}({\Theta}^y_*)-\cL^{s}({\Theta}_t)\big)dt,  
\end{align*}
where we use convexity of $\cL^{y}$ and that by convexity\footnote{In a convex function, for all $x$ and $y$ we have: $f(x)\geq f(y)+f'(y)(x-y)$.} of $\cL^{s}$ it holds $({\Theta}^y_*-{\Theta}_t)^\top\nabla_{{\Theta}}\cL^{s}({\Theta}_t)\leq \cL^{s}({\Theta}^y_*)-\cL^{s}({\Theta}_t)$. Integrating, we obtain
\begin{align*}
    V_t\leq e^{-t\beta^y\cL^{y}(\Theta_t)}V_0 + \beta^{s}\int_0^t e^{\beta^s\cL^{y}(\Theta_t)(u-t)}\big(\cL^{s}({\Theta}^y_*)-\cL^{s}({\Theta}_u)\big)du.
\end{align*}
This result shows that exact convergence is only obtained if ${\Theta}^y_*={\Theta}^s_*$ since in this case, by the convexity of $\mathcal{L}^s$, the right-most term would have been upper-bounded by zero as $\Theta_*^y$ would have been the optimizer also for $\mathcal{L}^s$. Otherwise, we only converge to a neighborhood of the optimum ${\Theta}^y_*$, where the size of the neighborhood is governed by $\beta^{s}$ and the properties of $\cL^s$. 

As a summary, the trade-off that we face for a \model with {\em limited} capacity is that: if we also optimize for the \priv attribute, we will only ever converge to a neighborhood of the optimum for the \pub attribute. While these results hold for a convex setting and are simplistic in nature, they give us a basic intuition into the idea that when the attributes $y$ and $s$ are somehow {\em correlated}, the output $\yh$ can better encode both tasks. This result also gives us the intuition that when we have {\em uncorrelated} attributes, we should need \models with more capacity to cover the payoff for not converging to the optimal point of \pub attribute.

\section{Pseudocode of Parameterized Attack}\label{appx:algo}
In Algorithm~\ref{alg:fg_ib} we show the pseudo-code of the training process we presented in Section~\textcolor{blue}{\ref{sec:IB}}. 
\begin{algorithm}[t]
\caption{Training an \hbc \model and its corresponding  attack  based on information bottleneck formulation in Section~\textcolor{blue}{\ref{sec:IB}}}\label{alg:fg_ib}
\begin{algorithmic}[1]
\STATE {\bfseries Input:} $\cF$: the model chosen as the classifier, $\cG$: the model chosen as the  attack , $\bbD^{train}$: training dataset, $(\beta^{x}, \beta^{y}, \beta^{s})$: trade-off multipliers, $E$: number of training epochs, $K$: batch size.  
\STATE {\bfseries Output:} Updated $\cF$ and $\cG$.
\STATE Random initialization of $\cF$ and $\cG$.
\FOR{$e: 1,\ldots, E$} 
\FOR{$b: 1,\ldots, |\bbD^{train}|/K$} 
\STATE $(\bX,\bY, \bS)$ $\leftarrow$ a random batch of $K$ samples $(\bx, y, s)\sim\bbD^{train}$ 
\STATE $\bYh = \cF(\bX)$ 
\STATE $\bSh = \cG(\bYh)$ 
\STATE $\hat{\tH}(S|\bYh) = - \sum_{k=1}^{K} \sum_{i=0}^{S-1} \bbI_{(s^k=i)} log (\sh^k_i)$ 
\STATE Update $\cG$ via the gradients of loss function $\cL^{\cG} = \hat{\tH}(S|\bYh)$
\STATE $\bSh = \cG(\bYh)$ 
\STATE $\hat{\tH}(S|\bYh) = - \sum_{k=1}^{K} \sum_{i=0}^{S-1} \bbI_{(s^k=i)} log (\sh^k_i)$ 
\STATE $\hat{\tH}(Y|\bYh) = - \sum_{k=1}^{K} \sum_{i=0}^{Y-1} \bbI_{(y^k=i)} log (\yh^k_i)$ 
\STATE $\hat{\tH}(\bYh) = - \sum_{k=1}^{K} \sum_{i=0}^{Y-1} \yh^k_i log (\yh^k_i)$  
\STATE Update $\cF$ via the gradients of loss function $\cL^{\cF} = \beta^{x}\hat{\tH}(\byh) + \beta^{y}\hat{\tH}(y|\byh) +  \beta^{s}\hat{\tH}(s|\byh)$
\ENDFOR
\ENDFOR
\end{algorithmic}
\end{algorithm}

\section{Additional Experimental Results}\label{appx:add_res}

\begin{table}[t]
\caption{The effect of \model's capacity on the UTKFace dataset for the \pub attribute Age and the \priv attribute Race where $Y=S=3$. Similar to Table~\textcolor{blue}{\ref{tab:entropy_betax_utk_33}}, we also show the average of the entropy of \model's output by  $\widetilde{\tH}(\byh)$ in bits. We fix ($\beta^x, \beta^y, \beta^s) = (.0, .7, .3)$. Model capacity is derermined in the first column, where a ``Layer Size'' of A-B-C-D-E refers to the number of neurons in each layer, respectively (related to the ``Output Size'' in Table~\textcolor{blue}{\ref{tab:dnn_architecture_F}}). We show results in two setting: (A) \HBCR and (B) \HBCP.} 
\label{tab:model_cap_utk_33}
\begin{tabular}{@{}lccc@{}}
Layers' Size & $\delta^y$     & $\delta^s$     & $\widetilde{\tH}(\byh)$ \\ \toprule
\multicolumn{4}{c}{(A) \HBCR}\\\midrule
128-128-64-64-128            & 81.17$\pm$0.27  &  83.14$\pm$0.48  &  0.61$\pm$0.03 \\
64-64-32-32-64             &  80.33$\pm$0.20  &  81.52$\pm$0.12  &  0.73$\pm$0.03 \\
42-42-21-21-42             &  78.45$\pm$0.79  &  79.50$\pm$0.38  &  0.86$\pm$0.03 \\
32-32-16-16-32             &   78.05$\pm$0.62  &  77.82$\pm$0.33  &  0.86$\pm$0.03 \\
25-25-12-12-25             &   75.73$\pm$0.61  &  74.96$\pm$0.80  &  1.00$\pm$0.11 \\
21-21-10-10-21             &   74.59$\pm$0.97  &  74.01$\pm$0.60  &  0.98$\pm$0.11 \\
18-18-9-9-18             &   73.94$\pm$0.63  &  71.39$\pm$0.72  &  1.12$\pm$0.02 \\
16-16-8-8-16             &   70.80$\pm$0.96  &  69.75$\pm$0.90  &  1.14$\pm$0.06 \\
12-12-6-6-12             &   64.68$\pm$5.02  &  66.39$\pm$2.65  &  1.31$\pm$0.05 \\
10-10-5-5-10             &   62.17$\pm$2.51  &  63.77$\pm$0.87  &  1.40$\pm$0.05 \\
8-8-4-4-8               &   62.74$\pm$2.97  &  63.20$\pm$1.20  &  1.41$\pm$0.03 \\ \midrule
\multicolumn{4}{c}{(B) \HBCP}\\\midrule
128-128-64-64-128            & 80.66$\pm$0.28 & 73.08$\pm$0.63 & 0.73$\pm$0.02           \\
64-64-32-32-64             & 79.96$\pm$0.32 & 72.67$\pm$0.20 & 0.80$\pm$0.03           \\
42-42-21-21-42             & 79.07$\pm$0.40 & 71.42$\pm$0.63 & 0.83$\pm$0.03           \\
32-32-16-16-32             & 77.29$\pm$0.43 & 69.87$\pm$1.04 & 0.88$\pm$0.02           \\
25-25-12-12-25             & 75.95$\pm$0.13 & 68.30$\pm$1.06 & 0.95$\pm$0.02           \\
21-21-10-10-21             & 75.00$\pm$0.68 & 67.41$\pm$1.04 & 1.00$\pm$0.04           \\
18-18-9-9-18             & 74.22$\pm$0.70 & 66.71$\pm$1.25 & 1.08$\pm$0.06           \\
16-16-8-8-16             & 73.53$\pm$0.42 & 67.89$\pm$0.87 & 1.21$\pm$0.05           \\
12-12-6-6-12             & 70.55$\pm$0.48 & 65.98$\pm$0.66 & 1.26$\pm$0.06           \\
10-10-5-5-10             & 66.46$\pm$2.59 & 63.24$\pm$1.17 & 1.30$\pm$0.05           \\
8-8-4-4-8               & 63.97$\pm$1.49 & 61.72$\pm$2.35 & 1.36$\pm$0.05           \\ \bottomrule
\end{tabular}
\end{table}  

In Table~\textcolor{blue}{\ref{tab:model_cap_utk_33}} we investigate the effect of the \model's capacity on its performance and behavior. Results show that both honesty ($\delta^y$) and curiosity ($\delta^s$) decrease by reducing the capacity of \model in a similar way. A more interesting observation is that the average of the outputs' entropy tends to increase when reducing the \model's capacity, which is due to both having lower honesty and also the emerged difficulties because of curiosity. This gives us the hint that by having a higher-capacity \model, a \server cannot only achieve better honesty-curiosity trade-offs, but also can easier hide the suspicious behavior of the \model's output.

Table~\textcolor{blue}{\ref{tab:race_age_utk}} and Table~\textcolor{blue}{\ref{tab:race_gender_utk}} show experimental results for settings where \pub and \priv attribute are reversed, compared to Table~\textcolor{blue}{\ref{tab:age_race_utk}} and Table~\textcolor{blue}{\ref{tab:gender_race_utk}} in Section~\textcolor{blue}{\ref{sec:eval}}. We can observe similar patterns in these results as well, confirming our findings explained in Section~\textcolor{blue}{\ref{sec:eval}}.

\begin{table*}[t]
\caption{The honesty  ($\delta^y$) and curiosity  ($\delta^s$) of an \hbc \model trained via regularized attack, where \pub attribute ($y$) is Race and \priv attribute ($s$) is Age. Values are in percentage (\%).}
\label{tab:race_age_utk} 
\resizebox{\textwidth}{!}{
\begin{tabular}{ccccc|cc|cc|cc} 
 & &  & \multicolumn{2}{c}{$S=2$} & \multicolumn{2}{c}{$S=3$} & \multicolumn{2}{c}{$S=4$} & \multicolumn{2}{c}{$S=5$} \\\cline{2-11}
          &    &   & $\delta^y$   & $\delta^s$   & $\delta^y$   & $\delta^s$   & $\delta^y$   & $\delta^s$   & $\delta^y$   & $\delta^s$   \\\cline{3-11}
\multirow{3}{*}{$Y=2$} & $\mathsf{NC}$ & (1.,\textit{0.}) & $87.63\pm.19$ & $\mathit{62.37\pm.41}$ & $87.63\pm.19$ & $\mathit{51.47\pm.28}$ & $87.63\pm.19$ & $\mathit{37.80\pm.56}$ & $87.63\pm.19$ & $\mathit{32.26\pm.07}$ \\
& \multirow{1}{*}{\HBCR} & (.7,.3) & $87.70\pm.26$  &  $81.42\pm.86$  &   $87.82\pm.19$  &  $79.58\pm.84$
  &  $88.17\pm.12$  &  $65.67\pm.79$ &  $87.94\pm.18$  &  $58.15\pm1.6$ \\ 
                   & \multirow{2}{*}{\HBCP} & (.7,.3) & $87.22\pm.19$  &  $67.22\pm.40$ &  $87.57\pm.33$  &  $54.03\pm.78$  &  $87.26\pm.16$  &  $40.86\pm.22$  & $87.24\pm.34$  &  $36.60\pm2.5$ \\  
                   &  & (.5,.5) & $86.58\pm.75$ & $71.78\pm.17$ & $82.97\pm1.6$ & $65.53\pm1.0$  & $80.67\pm1.2$ & $48.44\pm.87$ & $83.41\pm.31$ &  $53.96\pm1.3$ \\ \cline{3-11} 
\multirow{3}{*}{$Y=3$} & $\mathsf{NC}$ & (1.,\textit{0.}) & $85.42\pm.46$ & $\mathit{62.25\pm.87}$ & $85.42\pm.46$ & $\mathit{51.97\pm.43}$ & $85.42\pm.46$ & $\mathit{38.09\pm.77}$ & $85.42\pm.46$ & $\mathit{32.34\pm.45}$ \\ 
& \multirow{1}{*}{\HBCR} & (.7,.3) & $86.11\pm.42$  &  $82.14\pm.35$ &   $86.13\pm.31$  &  $78.09\pm.88$  &  $86.33\pm.08$  &  $65.56\pm.94$  &    $86.11\pm.34$  &  $58.95\pm1.3$    \\
                   & \multirow{2}{*}{\HBCP} & (.7,.3) & $85.25\pm.26$  &  $72.47\pm2.4$  &  $85.36\pm.61$  &  $64.52\pm1.0$  &   $85.31\pm.38$  &  $53.03\pm.90$ & $85.19\pm.10$  &  $45.77\pm.51$ \\
                   &  & (.5,.5) & $84.12\pm.27$ & $81.16\pm.23$  & $76.85\pm.38$ & $79.74\pm.64$ & $76.00\pm.31$ & $66.05\pm.46$ & $75.37\pm.27$ & $60.10\pm.49$ \\ \cline{3-11} 
\multirow{3}{*}{$Y=4$} & $\mathsf{NC}$ & (1.,\textit{0.}) & $81.57\pm.33$ & $\mathit{64.99\pm.70}$ & $81.57\pm.33$ & $\mathit{54.12\pm.58}$ & $81.57\pm.33$ & $\mathit{41.13\pm.15}$ & $81.57\pm.33$ & $\mathit{35.56\pm.47}$ \\
& \multirow{1}{*}{\HBCR} & (.7,.3) & $81.69\pm.12$  &  $81.33\pm.38$  &  $81.08\pm.57$  &  $77.57\pm.54$   &   $81.26\pm.35$  &  $64.44\pm.54$  &     $81.15\pm.22$  &  $58.93\pm.82$ \\
                   & \multirow{2}{*}{\HBCP} & (.7,.3) & $81.20\pm.21$  &  $75.17\pm.35$  &  $81.27\pm.48$  &  $69.14\pm.07$  &   $81.09\pm.10$  &  $55.63\pm.11$  & $81.21\pm.08$  &  $51.28\pm.52$ \\
                   &  & (.5,.5) & $78.09\pm.12$ & $81.02\pm.58$ & $71.89\pm.27$ & $78.72\pm.39$  & $70.43\pm.61$ & $65.99\pm.75$ & $69.81\pm.45$ & $58.89\pm.37$ \\ \cline{3-11} 
\multirow{3}{*}{$Y=5$} & $\mathsf{NC}$ & (1.,\textit{0.}) & $80.66\pm.24$ & $\mathit{66.29\pm.39}$ & $80.66\pm.24$ & $\mathit{56.64\pm.25}$ & $80.66\pm.24$ & $\mathit{44.42\pm.76}$ & $80.66\pm.24$ & $\mathit{38.13\pm.73}$ \\ 
& \multirow{1}{*}{\HBCR} & (.7,.3) & $81.05\pm.40$  &  $80.93\pm.78$ &   $80.78\pm.66$  &  $78.48\pm.26$  &  $80.00\pm.45$  &  $65.16\pm.46$  &      $80.87\pm.70$  &  $58.71\pm.24$ \\   
                   & \multirow{2}{*}{\HBCP} & (.7,.3) & $80.94\pm.35$ & $79.59\pm.27$ &  $80.59\pm.08$  &  $76.57\pm.28$  &   $80.37\pm.09$  &  $62.76\pm.47$  &  $80.52\pm.06$  &  $55.75\pm.52$ \\
                   &  & (.5,.5) & $80.03\pm.22$ & $81.46\pm.23$  & $72.36\pm.55$ & $78.57\pm.17$ & $70.48\pm1.6$ & $66.32\pm.41$ & $70.88\pm.66$ & $58.02\pm1.2$ \\\bottomrule 
\end{tabular}%
}
\end{table*}
\begin{table*}[t]
\centering
\caption{{\bf Gender and Race.} Classification accuracy~(\%) on UTKFace test set for Race vs. Gender. In each experiment, there are 4 different tasks based on the size $Y$ and $S$. Empty cells (---) are due to the incapability of regularized attacks for $S>2$. }
\label{tab:race_gender_utk}
\resizebox{\textwidth}{!}{%
\begin{tabular}{cccccc|cc|cc|cc}
 & & & & \multicolumn{2}{c}{$S=2$} & \multicolumn{2}{c}{$S=3$} & \multicolumn{2}{c}{$S=4$} & \multicolumn{2}{c}{$S=5$} \\\cline{5-12}
& Setting &  Attack  & $(\beta^y, \beta^s)$ & $\delta^y$   & $\delta^s$   & $\delta^y$   & $\delta^s$   & $\delta^y$   & $\delta^s$   & $\delta^y$   & $\delta^s$   \\\cline{2-12}
\multirow{6}{*}{$Y=2$} & \NC & \Rval\cite{song2019overlearning} & (1., \textit{0.}) & $90.23\pm.21$ &  $\mathit{58.62\pm1.2}$ & $90.23\pm.21$ & $\mathit{47.54\pm1.1}$ & $90.23\pm.21$ & $\mathit{44.21\pm.43}$ & $90.23\pm.21$ & $\mathit{43.94\pm.04}$ \\\cline{3-12}  
                   & \HBCR & Parameterized & (.7, .3) & $90.18\pm.11$ & $86.91\pm.32$ & $90.56\pm.23$ &  $81.66\pm.24$ & $90.24\pm.09$ & $74.04\pm.48$ & $90.28\pm.35$ & $74.01\pm.89$  \\\cline{3-12}
                   & \multirow{5}{*}{\HBCP} & \multirow{2}{*}{Parameterized} & (.7, .3)  & $90.59\pm.07$ & $56.21\pm.00$  & $89.92\pm.11$ & $46.26\pm.50$ &  $90.37\pm.19$ & $44.31\pm.21$ & $90.14\pm.34$ & $44.93\pm.94$ \\ 
                   &  & & (.5, .5) & $89.74\pm.43$ & $61.60\pm4.1$ & $86.95\pm.66$ & $52.85\pm2.3$ & $87.69\pm.48$ & $50.57\pm.53$ & $87.57\pm.76$ & $50.88\pm.78$  \\
                   &  & & (.3, .7) & $79.96\pm.80$ & $81.54\pm.60$ & $72.41\pm.55$ & $77.28\pm.45$ & $72.97\pm7.3$ & $63.90\pm2.6$ & $72.97\pm7.3$ & $63.90\pm2.6$  \\\cline{4-12}   
                   &&\multirow{3}{*}{Regularized} & (.7, .3) & $90.18\pm.17$  &   $73.20\pm.10$ &---&---&---&---&---&---\\
                   &&& (.5, .5) & $89.71\pm.06$  &  $82.28\pm.10$ &---&---&---&---&---&---\\ 
                   &&& (.3, .7) & $87.86\pm.40$  &  $84.35\pm.48$ &---&---&---&---&---&---\\\bottomrule 
\end{tabular}%
}
\end{table*}


\section{Curiosity and Generalization}\label{sec:generalization} 
Underlying the capabilities of the \hbc \models is the ability of the output $\byh$ to encode multiple attributes. Such capability also forms the foundation for {\em representation learning}~\cite{bengio2013representation}, {\em multi-task learning}~\cite{finn2017model} and {\em disentanglement of factors}~\cite{marx2019disentangling}.  It is well-known that added noise during training, \eg in the form of Dropout, improves the generalization performance of a model \cite{vera2018role}. These observations coincide with results obtained in~\cite{melis2019exploiting}, where it is shown that differential privacy is not effective protection against property inference attacks; even more so, noise may improve the performance by keeping the model from overfitting on the main task. 

Information bottleneck~(IB) is used for understanding the nature of intermediate representations produced by the hidden layers of DNNs during training~\cite{tishby2015deep, shwartz2017opening, saxe2019information}. In Section~\textcolor{blue}{\ref{sec:IB}}, we discussed that  formulation in Eq~\eqref{eq:optim_F_1} shows a trade-off: compressing the information included in $\bx$, while encoding as much information about $y$ and $s$ as possible into $\byh$. In other words, IB states that the optimal \hbc \model must keep the information in $\bx$ that is useful for $y$ and $s$ while compressing other, irrelevant information. On the other hand, a useful \model is one that does not {overfit} on its training dataset, meaning that the {\em generalization gap}\footnote{The difference between the \model's performance on its training data and the \model's performance on unseen test data is called the generalization gap of \model.} is small. 

One can notice that under sufficient ``similarity'' between the \pub and \priv attribute, the addition of the \priv attribute into the \model's optimization objective can even {\em improve} the generalization capabilities for the classification of \pub attribute~\cite{lee2019generalization}. Specifically, the addition of another attribute $s$ while training for the \pub attribute $y$ can reduce the generalization gap as the addition of another attribute can act as a {\em regularizer}, which places an inductive bias on the learning of the \pub attribute and guides the model towards learning more general and discerning features~\cite{caruana1997multitask,zhang2020generalization}. 

Previous work (\eg Theorem 1 in~\cite{vera2018role}) has shown that the generalization gap can be upper-bounded by the amount of compression happening in the \model, \ie the mutual information between the internal representations learned by the \model and the input data. Basically, the compression forces the \model to throw away information unrelated to the \pub task~(\eg noises). Moreover, \cite{lee2019generalization} shows that training a \model for more than one attribute can bring a major benefit in situations where the single-attribute \model ``underperforms'' on the \pub task, \eg due to the lack of samples in the training data or the presence of noise. Particularly this happens when attributes are related---where the relatedness is measured through the alignment of the input features---and the data for the second attribute is of good quality~\cite{lee2019generalization}. 

Our observation from Section~\textcolor{blue}{\ref{sec:eval}} is that, while properties Age, Race, and Gender are not necessarily correlated properties, they are all coarse-grained properties for which understanding of the full image is required (as opposed to fine-grained properties present only in part of the image). As observed from the results, in certain settings, including curiosity in the classifier can improve the performance of the main task. Therefore, because all properties are sufficiently coarse-grained, learning in a multi-task framework can keep the network from overfitting details by enforcing it to learn certain general, coarse-grained features and thus perform better in the main task.

Putting all together, it can be expected that by training an \hbc \model, we can not only reduce the generalization gap for the \pub attribute, but also compress the learned representations, which means decrease~(or at least not increase) the entropy of the representations. Thus, in situations where the curiosity of the \model {\em improves} generalization, the recognition of an \hbc \model is even more challenging. 

\begin{figure}[t!]
    \centering
    \includegraphics[width=.95\columnwidth]{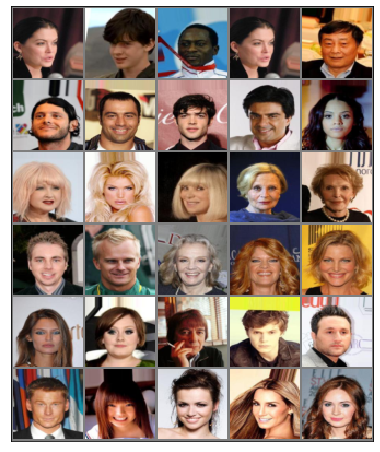}
    \caption{Sample images from CelebA~\cite{liu2015celeba}. Columns (from left to right) show Male, Smiling, Attractive, High Cheekbones, and Heavy Makeup attributes. The first and second rows show Black Hair with and without the other corresponding column's attribute, respectively. Similarly, the third and fourth rows Blond Hair, and fifth and sixth rows Brown Hair attributes.}
    \label{fig:sample_imgs_celeba}
\end{figure}
\begin{figure}[t!]
    \centering
    \includegraphics[width=.95\columnwidth]{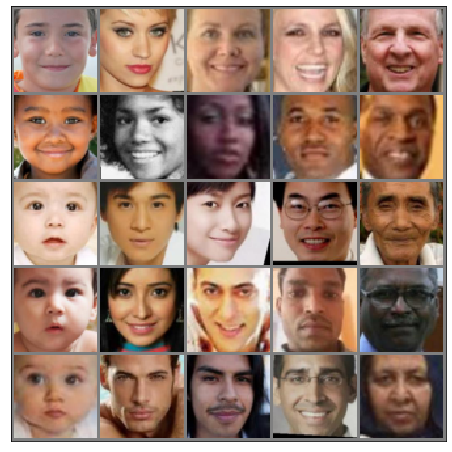}
    \caption{Sample images from UTKFace~\cite{zhang2017age}. Each row shows images labeled with the same Race (White, Black, Asian, Indian, and others), and each column shows one of the five Age groups ([0-19,20-26,27-34,35-49, 50-100].}
    \label{fig:sample_imgs_utk}
\end{figure}

\section{More Motivational Examples}\label{sec:more_motiv_example}
As some other motivational examples, we can consider:\\

(1) A sentiment analysis application that can be run on the \user's browser and, taking a text, it can help the user to estimate how their text may sound to the readers: \eg $\byh = [p_{positive}, p_{neutral}, p_{negative}]$. The \server may ask for collecting only outputs, \eg to improve their service, but \server might try to infer if the text includes specific \priv words or if the text falls into a certain category; by making an \hbc sentiment \model. 

(2) An activity recognition app, from an insurance company, can be run on the \user's smartwatch and analyzing motion data. This app can provide an output $\byh=[p_{walk}, p_{run}, p_{sit}, p_{sleep}]$. The \server may ask for only collecting $\byh$ to offer the \user further health statistics or advice, but it may also try to infer if the \user's body mass index (BMI) is more than or below a threshold. 

(3) Automatic speech recognition~(ASR) is the core technology for all voice assistants, such as Amazon Alexa, Apple Siri, Google Assistant, Microsoft Cortana, to provide specialized services to their users, \eg playing a song, reporting weather condition, or ordering food. ASR models, at the top, have a \texttt{softmax} layer that estimates a probability distribution on a vocabulary with a particular length, \eg 26 letters in the English alphabet plus some punctuation symbols. Service providers usually perform post-processing on the output of the ASR model, that is the probable letter for each time frame, to combine them and predict the most probable word and further the actual sentence uttered by the \user. Despite the fact that they can be asked to run their ASR model at the \user's side, but they can argue that the post-processing ASR's outputs is crucial for providing a good service, and these outputs are supposed to only predict what the \user is asking for (\ie the uttered sentence) and not any other information, \eg the Age, Gender, or Race of the \user. However, an \hbc ASR might secretly violate its user's privacy.

(4) An EU-funded project~\cite{sherpa2021} designed an app~\cite{hownorm2021} for people to experience how ML models judge their faces. This app asked for access to the \user's camera to capture their face image. Then, pre-trained ML \models are being run on the \user's own computer, in the browser. The project promises \users that ``no personal data is collected'' but at the end, \users can voluntarily ``share some anonymized data''~\cite{hownorm2021}. Despite the useful purpose of this app, we show that even such anonymous collection of the outputs of pre-trained \models can result in some privacy leakage.

\section{Datasets}

\subsection{Details of CelebA Dataset}\label{apx:celeba}

Figure~\textcolor{blue}{\ref{fig:sample_imgs_celeba}} shows some sample images from CelebA dataset~\cite{liu2015celeba}. While CelebA does not have any attributes with a number of classes more than $2$, we utilize three mutually exclusive binary attributes BlackHair, BlondHair, BrownHair to build a 3-class attribute of HairColor. This still gives us a dataset of more than 115K images that fall into one of these categories. 
 
\subsection{Details of UTKFace Dataset}\label{apx:utk}
Figure~\textcolor{blue}{\ref{fig:sample_imgs_utk}} shows some sample images from UTKFace dataset~\cite{zhang2017age}.
Each image, collected from the Internet, has three attributes: (1)~Gender (male or female), (2)~Race (White, Black, Asian, Indian, or others), and (3)~Age (from 0 to 116).  Gender is a binary label and there are 12391 vs. 11314 images with male vs. female labels. Table~\textcolor{blue}{\ref{tab:apx_utk_ages}} and Table~\textcolor{blue}{\ref{tab:apx_utk_races}} show how we assign Age and Race labels to each image in UTKFace for different number of classes used in our evaluations. For the Age label we choose categories such that classes become balanced, in terms of the number of samples. For the Race label, we are not that free (like Age), thus we choose categories such that classes do not become highly unbalanced and we try to keep samples in each class as similar to each other as possible. 
\begin{table}[t]
\caption{The details of how we assign  {\em Age} label to each image in UTKFace.}
\label{tab:apx_utk_ages}
\resizebox{\columnwidth}{!}{%
\begin{tabular}{cccccl}
 & \multicolumn{4}{c}{Number of Classes} \\
Label & 2 & 3 & 4 & 5 & Samples\\ \cline{2-5} 
0 & a $\leq$ 30 & a $\leq$ 20 & a $\leq$ 21 & a $\leq$ 19 & 4593\\
1 & 30 $>$ a & 20  $<$  a $\leq$ 35 & 21 $<$ a $\leq$ 29 & 19 $<$ a $\leq$ 26 & 5241\\
2 &  & 35 $<$ a & 29 $<$ a  $\leq$ 45 & 26 $<$ a $\leq$ 34 & 4393\\
3 &  &  & 45 $<$ a & 34 $<$ a $\leq$ 49 & 4491 \\
4 &  &  &  & 49 $<$ a & 4987\\\bottomrule
\end{tabular}%
}
\end{table}
\begin{table}[t]
\caption{The details of how we assign  {\em Race} label to each image in UTKFace. W:White, B:Black, A:Asian, I:Indian, or O:others.}
\label{tab:apx_utk_races}
\begin{tabular}{cccccl}
 & \multicolumn{4}{c}{Number of Classes} &\\
Label & 2 & 3 & 4 & 5 &  Samples \\ \cline{2-5} 
0 & W & W  & W & W & 10078\\
1 & BAIO & A & B & B & 4526\\
2 &  & BIO & I & A & 3434\\
3 &  &  &  AO & I & 3975\\
4 &  &  &  & O & 1692\\\bottomrule
\end{tabular}%
\end{table}

\section{Model Architectures}\label{apx:dnn_arch}
For all experiments reported in this paper, we use PyTorch~\cite{paszke2019pytorch} that is an open-source library for the implementation of deep neural networks~\cite{lecun2015deep, schmidhuber2015deep}. Table~\textcolor{blue}{\ref{tab:dnn_architecture_F}} and Table~\textcolor{blue}{\ref{tab:dnn_architecture_G}} show the details of neural network architecture that are implemented as \model $\cF$ and  attack  $\cG$ (see Figure~\textcolor{blue}{\ref{fig:adv_train}}).

\begin{table}[t]
\caption{The implemented model as \model $\cF$.}
\label{tab:dnn_architecture_F}
\resizebox{\columnwidth}{!}{%
\begin{tabular}{@{}lcr@{}}
\toprule
Layer Type & Output Size & Number of Parameters \\ \midrule
Conv2d(kernel=2, stride=2) & 128, 32, 32 & 1,664 \\
BatchNorm2d & 128, 32, 32 & 256 \\
LeakyReLU(slope=0.01) & 128, 32, 32 & 0 \\
Dropout(p=0.2) & 128, 32, 32 & 0 \\
Conv2d(kernel=2, stride=2) & 128, 16, 16 & 65,664 \\
BatchNorm2d & 128, 16, 16 & 256 \\
LeakyReLU(slope=0.01) & 128, 16, 16 & 0 \\
Dropout(p=0.2) & 128, 16, 16 & 0 \\
Conv2d(kernel=2, stride=2) & 64, 8, 8 & 32,832 \\
BatchNorm2d & 64, 8, 8 & 128 \\
LeakyReLU(slope=0.01) & 64, 8, 8 & 0 \\
Dropout(p=0.2) & 64, 8, 8 & 0 \\
Conv2d(kernel=2, stride=2) & 64, 4, 4 & 16,448 \\
BatchNorm2d & 64, 4, 4 & 128 \\
LeakyReLU(slope=0.01) & 64, 4, 4 & 0 \\
Dropout(p=0.2) & 64, 4, 4 & 0 \\
Linear & 128 & 131,200 \\
LeakyReLU(slope=0.01) & 128 & 0 \\
Dropout(p=0.5) & 128 & 0 \\
Linear & Y & 128$\times$Y+Y \\
\bottomrule
\end{tabular}%
}
\end{table}
\begin{table}[t]
\caption{The implemented model as  attack  $\cG$.}
\label{tab:dnn_architecture_G} 
\resizebox{\columnwidth}{!}{%
\begin{tabular}{@{}lcr@{}}
\toprule
Layer Type & Output Size & Number of Parameters \\ \midrule
Linear & 20$\times$Y & 80$\times$Y \\
LeakyReLU(slope=0.01) & 20$\times$Y & 0 \\
Dropout(p=0.25) & 20$\times$Y & 0 \\
Linear & 10$\times$Y & 610$\times$Y \\
LeakyReLU(slope=0.01) & 10$\times$Y & 0 \\
Dropout(p=0.25) & 10$\times$Y & 0 \\
Linear & S & 10$\times$Y$\times$S+S \\ 
\bottomrule
\end{tabular}%
}
\end{table}

\end{document}